\documentclass{article}

\usepackage[preprint]{neurips_2026}

\usepackage[utf8]{inputenc} 
\DeclareUnicodeCharacter{2265}{\ensuremath{\geq}}
\usepackage[T1]{fontenc}    
\usepackage{url}            
\usepackage{booktabs}       
\usepackage{amsfonts}       
\usepackage{nicefrac}       
\usepackage{microtype}      
\usepackage{xcolor}         

\usepackage{graphicx}
\usepackage{subcaption}
\usepackage{fvextra}

\usepackage{amsmath}
\usepackage{amssymb}
\usepackage{mathtools}

\usepackage{natbib}

\usepackage[colorlinks=true, linkcolor=black, citecolor=blue, urlcolor=blue]{hyperref}

\usepackage[capitalize,noabbrev]{cleveref}
\DefineVerbatimEnvironment{PromptVerbatim}{Verbatim}{breaklines=true,breakanywhere=true,fontsize=\small}

\title{Emergent Manifold Separability during Reasoning in Large Language Models}

\author{%
  Chanwoo Chun\textsuperscript{1,2}\thanks{Equal contribution.} \quad
  Alexandre Polo\textsuperscript{1}\footnotemark[1] \quad
  SueYeon Chung\textsuperscript{1,3,4}\\
  \textsuperscript{1}Department of Physics, Harvard University\\
  \textsuperscript{2}Center for Data Science, New York University\\
  \textsuperscript{3}Kempner Institute, Harvard University\\
  \textsuperscript{4}Center for Computational Neuroscience, Flatiron Institute\\
  \texttt{\{cchun, apolo, schung\}@fas.harvard.edu}
}

\begin{document}

\maketitle

\begin{abstract}
Chain-of-Thought (CoT) prompting significantly improves reasoning in Large Language Models, yet the temporal dynamics of the underlying representation geometry remain poorly understood. We investigate these dynamics by applying Manifold Capacity Theory (MCT) to two compositional reasoning tasks: a controlled Boolean logic tree that supports deep mechanistic analysis, and a natural-language eligibility task in which the model has to extract attributes from prose, compare them to thresholds, and compose the local decisions through a fixed evaluation tree. MCT lets us quantify the linear separability of latent representations without the confounding factors of probe training. On both tasks, and across several open-weight models, reasoning manifests as a transient geometric pulse: concept manifolds are untangled into linearly separable subspaces immediately prior to computation and rapidly compressed thereafter. This behavior diverges from standard linear probe accuracy, which remains high long after computation, suggesting a fundamental distinction between information that is merely retrievable and information that is geometrically prepared for processing. We interpret this phenomenon as \emph{Dynamic Manifold Management}, a mechanism where the model dynamically modulates representational capacity to optimize the bandwidth of the residual stream throughout the reasoning chain.
\end{abstract}


\section{Introduction}


Chain-of-Thought (CoT) prompting is highly effective at enabling Large Language Models (LLMs) to solve complex compositional problems \citep{wei2022chain}. The intuition is that CoT distributes the computational load: rather than relying solely on feed-forward depth, the model uses recursive token generation to process information over time. However, the precise mechanisms governing this temporal processing remain largely uncovered.

Studies of feed-forward computation have shown that layerwise representation geometry can provide critical insight into information flow \citep{cohen2020separability,pope2021intrinsic,raghu2017svcca}. Far less is known about how representation geometry evolves over tokens. We bridge this gap by borrowing geometric measures from computational neuroscience and deep neural networks \citep{barrett2019analyzing} to track how internal computation unfolds over time.

Existing research on CoT internal representations often focuses on whether the textual output faithfully reflects the model's hidden computation \citep{turpin2023faithful,lanham2023faithfulness}. Recent works have used linear probes to show that intermediate answers are often linearly decodable well before they are generated \citep{zhang2025reasoning,afzal2025knowing}.
However, standard probing techniques have inherent limitations that may lead to overestimating a model's actual knowledge \citep{afzal2025knowing}, and can even yield high accuracy when probes are trained on random features \citep{zhang2018language,wieting2019no}.

To address these limitations, we analyze the latent space using Manifold Capacity Theory (MCT) \citep{chung2018classification} alongside linear probing. Unlike probing, MCT quantifies linear separability as an intrinsic geometric property, eliminating the need to train a classifier. This framework has provided key insights into information processing in visual and speech systems and early language models \citep{chou2024geometry,stephenson2019untangling,stephenson2021geometry,mamou2020emergence,kirsanov2025geometry}.

\begin{figure}[t]
  \begin{center}
    \centerline{\includegraphics[width=0.7\columnwidth]{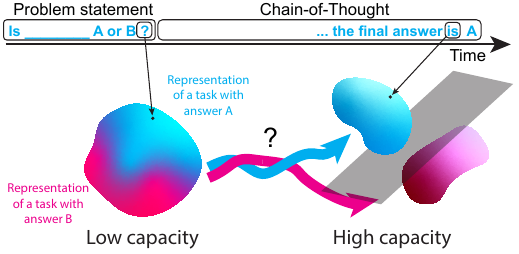}}
    \caption{
    \textbf{Manifold Untangling during Chain-of-Thought.}
    Before any token is generated (left) the latent representations of tasks corresponding to answers A and B are entangled and hard to distinguish resulting in low capacity.
    As the model generates the CoT, it progressively untangles these representations. By the final token (right), the two manifolds are linearly separable and capacity is high.
    }
    \vspace{-0.8cm}
    \label{fig:manifold_untangling}
  \end{center}
\end{figure}

In this work, we use MCT to map the dynamics of representation manifolds in the residual stream during compositional reasoning (\cref{fig:manifold_untangling}). We study two complementary tasks: a controlled Boolean logic tree, and a natural-language eligibility task in which the model must extract attributes from prose, compare them to thresholds, and compose the resulting decisions through a fixed evaluation tree. We additionally replicate the central effects across several open-weight models.

Our core contributions are:
\begin{enumerate}
    \item Geometric evidence, provided by Capacity Theory, that CoT involves active reasoning throughout generation, rather than only mimicking a reasoning trace.

    \item The LLM dynamically modulates the representation geometry such that only the concepts relevant for a computation at a given time have high manifold capacity. We find that this mechanism keeps the representation dimensionality low, saving the bandwidth budget of the residual stream.
\end{enumerate}

\section{Related Work}
\subsection{Internal Mechanisms of Chain-of-Thought}
While the performance benefits of CoT are well-established, the nature of the underlying internal computation remains debated. One line of inquiry questions the \emph{faithfulness} of the generated reasoning, suggesting that CoT may be a post-hoc rationalization rather than a faithful account of the computation that leads to the prediction \citep{turpin2023faithful,lanham2023faithfulness}. Smaller, distilled models may be particularly at risk of that, and it has been shown that for small models, more CoT granularity does not necessarily improve performance \citep{chen2025distill}.
On the mechanistic side, recent interpretability work has identified a circuit within a two-layer attention-only model that plays a role in inductive reasoning \citep{elhage2021mathematical,olsson2022inductionheads}.

A growing body of recent literature focuses specifically on decoding the reasoning states directly from the latent activations.
\cite{zhang2025reasoning} and \cite{afzal2025knowing} investigate the necessity of a long CoT and demonstrate that the correctness of an intermediate step is often linearly decodable long before the token is generated, suggesting that the model maintains a coherent internal truth state. This observation has also motivated early-stopping strategies for reasoning models.

Our work contributes to this discourse by looking for global geometric signatures of the reasoning steps, and specifically investigates the dynamics of the geometric representation along the CoT.

\subsection{Linear Representations and Probing Challenges}

The prevailing view in mechanistic interpretability is the \emph{Linear Representation Hypothesis}, which suggests that neural networks encode semantic concepts such as sentiment, syntax, or factual truth as vectors in a linear subspace \citep{elhage2022toymodels}. Under this hypothesis, linear probes, supervised classifiers trained on frozen activations, have become a popular tool. They have been used to decode various concepts, from syntactic tree depth \citep{hewitt2019structural} to the truthfulness of model statements \citep{azaria2023internalstatellmknows,burns2024discoveringlatentknowledgelanguage,marks2024geometrytruthemergentlinear}, and even spatiotemporal coordinates \citep{gurnee2024languagemodelsrepresentspace}, serving as a powerful diagnostic tool for locating information within the residual stream.

However, despite their empirical success, relying solely on probing accuracy presents significant methodological challenges. First, the probe is sensitive to regularization, optimizer, early stopping, class
imbalance handling, and train/test splits. Moreover,
\cite{zhang2018language}, \cite{wieting2019no}, and \cite{hewitt2019designing} established that linear probes can achieve high accuracy on random control tasks by exploiting the high-dimensional nature of the representation space or the structural priors of the model architecture. This has prompted the need to find a more geometric approach to interpretability research \citep{afzal2025knowing}.

\subsection{Manifold Theory and Representational Geometry}
Geometry-based analysis techniques have been actively developed and widely used for studying the brain and feed-forward networks \citep{pope2021intrinsic, lehky2014dimensionality,willett2021high,ansuini2019intrinsic,morcos2018insights}. This approach treats a set of neural representations of inputs as a single entity: a neural manifold. The perspective that neural computation aims to disentangle these manifolds (e.g., manifolds of dog images and cat images being disentangled for a linear readout, see \cref{fig:manifold_untangling} for a schematic) has provided critical insights into the brain \citep{dicarlo2007,yamins2016using}. This geometric perspective has been formalized through Manifold Capacity Theory \citep{chung2018classification, cohen2020separability,chou2024geometry} and has provided useful insights into feedforward computation in the visual and auditory cortex \citep{chou2024geometry} and in artificial vision models, where class manifolds are observed to progressively \emph{untangle} over layers \citep{stephenson2019untangling}. Its application to Large Language Models remains nascent: \cite{mamou2020emergence} analyzed the geometry of static word embeddings and recent work has begun to explore the intrinsic dimension of transformer representations \citep{ansuini2019intrinsic,kirsanov2025geometry}. We extend this lineage by applying Manifold Capacity to the temporal dimension, tracking the dynamic modulation of concept manifolds during reasoning.

\section{Tasks}

We study two compositional reasoning tasks that share the same abstract tree structure. The Boolean tree task is parametric, so it is predictably controllable, which lets us run the deepest analyses on it. The eligibility task is a more natural extension where the same tree-structured reasoning has to be carried out on top of a prose description.

\subsection{Compositional Boolean Logic Task}
As a representative benchmark for tasks requiring multi-step reasoning, we utilize a hierarchical Boolean logic task. The goal is to evaluate the Boolean value of a nested expression, such as \texttt{((True and False) or (True xor True))}. Each expression corresponds to a full binary tree of height $h$, where internal nodes represent logical operators \verb|AND, OR, XOR| and leaf nodes provide Boolean constants.

To facilitate a structured analysis of the internal representations, we employ a consistent naming and ordering scheme. Each internal node is assigned a unique identifier \verb|[ID]| based on a post-order or level-order traversal of the logic tree. This ensures that, with an appropriate prompt strategy, all tasks are solved in the same order. For our primary analysis on the internal representation, we generate a balanced dataset of 256 expressions (128 True, 128 False at the root) for a tree of height $h=5$, containing $2^h - 1 = 31$ internal nodes.

We use Ministral 3 8B Reasoning \citep{ministral3} as the primary model on this task, and additionally replicate the main results on Qwen2.5-14B-Instruct and gpt-oss-20b.

\begin{figure}[t]
  \begin{center}
    \centerline{\includegraphics[width=0.8\columnwidth]{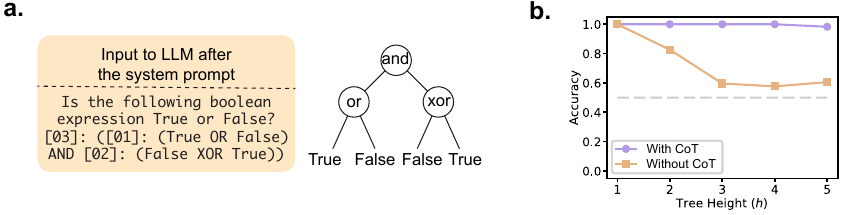}}
    \caption{
    \textbf{a.} Example of a Boolean logic tree with height $h=2$. \emph{Left}: The corresponding text input provided to the model, where internal nodes are labeled [01] through [03]. \emph{Right}: Schematic of the tree structure.
    \textbf{b.} Accuracy on Boolean logic trees of varying height. Performance with CoT (purple) remains near-perfect ($>98\%$), while standard prompting (No CoT, orange) degrades significantly as tree depth increases ($59\%$ for $h=5$). The dashed line marks the random baseline (50\%).
    }
    \vspace{-1cm}
    \label{fig:tree_accuracy}
  \end{center}
\end{figure}

\subsection{Eligibility Task}\label{sec:eligibility_task}

The eligibility task replaces the symbolic Boolean leaves with natural-language facts that have to be extracted and compared. Each item gives the model a short applicant paragraph describing attributes such as age, income, credit score, and savings, together with a list of eligibility criteria and group rules organized as a small fixed tree. The model has to read the paragraph, recover the relevant facts in the presence of distractor attributes, compare each fact against its threshold, and then compose the local decisions through the tree. The shared evaluation order is $C_1 \to C_2 \to G_1 \to C_3 \to C_4 \to G_2 \to \texttt{Final}$, where each $C_i$ is a leaf criterion (e.g., $\text{age} \ge 40$), $G_1$ and $G_2$ are group nodes that combine two criteria with \texttt{AND} or \texttt{OR}, and \texttt{Final} combines the two groups. A concrete example input and a representative model response are provided in \cref{sec:app_eligibility_example}.

The dataset is generated by a script that first samples the desired ground-truth labels for all criteria, group nodes, and the final node, then samples concrete thresholds and applicant values that realize those labels, and finally renders them into a prose paragraph using varied templates, shuffled sentence order, and distractor attributes. This yields 256 items with known intermediate and final labels by construction, while still requiring genuine extraction and comparison from the model.

We use Ministral 3 8B Reasoning as the primary model on this task as well, and additionally replicate the unstructured variant on Qwen2.5-7B-Instruct.

\section{Methods}

\subsection{Prompting Strategy and Structural Anchoring}\label{sec:prompting}

To capture the evolution of neural manifolds, we ask the model to follow a structured Markdown output format with consistent phase markers. For the Boolean task, the model produces a \textit{Problem Statement}, then a \textit{Solve} phase that walks through one node at a time using \texttt{**Node [k]**}, \texttt{* Logic:}, and \texttt{* Result:} lines, and finally a \textit{Summary} that recalls all intermediate results. The eligibility task uses a structurally analogous template (\texttt{Extract / Threshold / Compare / Result} for each criterion, \texttt{Logic / Result} for each group, then a \texttt{Final} step). The full system prompts and a representative model response for each task are provided in \cref{sec:app_prompts,sec:app_eligibility_example}. With these prompts, our primary model achieves $98\%$ all-node accuracy on depth-5 Boolean trees and perfect final and all-node accuracy on the 256-example standard eligibility run (see \cref{sec:task_perf}).

The fixed format gives us deterministic textual markers that we use as \emph{temporal anchors} to align internal representations across items, in the spirit of recent practice on probing the intermediate states of reasoning models \citep{zhang2025reasoning,afzal2025knowing}. Concretely, each anchor is a short formatting string that marks a consistent point in a node's local computation, such as the node header, the \texttt{* Logic:} line, or the colon at the end of \texttt{* Result:} on the Boolean task; for the eligibility task, the analogous anchors include criterion/group headers, result lines, and the summary markers such as \texttt{C1:}. We compute manifold capacity and probe accuracy at the \emph{final token} of each anchor string. For result anchors, this token sits before any textual leakage of the corresponding answer; for the other anchors, it provides a fixed phase boundary before the model writes the next part of the node computation. These anchors provide natural alignment points across items that otherwise generate different numbers of tokens. To minimize formatting variance, we set $T = 0$ and disable sampling, and use the eager-attention implementation for stable activation capture. Experiments are run on NVIDIA A100 (80GB) and H100 GPUs.

\subsection{Analysis Methods}
To solve a given task, the model has to successively determine the label of each node. At any structural anchor and layer, we group activations by the ground-truth label of a chosen node and quantify how strongly that node's information is represented. We do this either with manifold capacity theory or with a trained linear probe. The tokens along the CoT generation that we analyze are the structural anchors described in \cref{sec:prompting}, and unless stated otherwise the activations are taken from the post-MLP residual stream.

\paragraph{Manifold Capacity}
We measure the manifold capacity on a set $\mathcal{M}\coloneq \{\mathbf{x}_i\}_{i=1}^{D}$ of intermediate-layer residual-stream embeddings aligned at a single structural anchor across $D$ items, with binary labels $\mathcal{Y}\coloneq \{y_i \}_{i=1}^{D}$ given by the ground-truth label of a single node in the underlying tree (e.g.\ True/False on Boolean, Met/Not Met on eligibility).

MCT gives $N^{*}$, the minimal random subspace dimension needed for linear separation under a given labeling, which can be efficiently computed following \cite{chou2024geometry}:
\[
N^{*}=\left\langle \|\Pi_{\mathcal{V}}(\mathbf{t})\|_{2}^{2}\right\rangle _{\mathbf{t}\sim\mathcal{N}\left(0,\mathbf{I}_{N}\right)},
\]
where $\Pi_{\mathcal{V}}(\mathbf{t})$ is a Euclidean projection of
the $\mathbf{t}$ vector to a cone $\mathcal{V}$ of the
points $\left\{ y_{i}\mathbf{x}_{i}\right\} _{i=1}^{D}$, with $y_{i}\in\{+1,-1\}$ encoding the binary label of the chosen node. In our binary-label setting, $P=2$ denotes the two class manifolds, so we compute the capacity as $\alpha=P/N^*=2/N^*$. Because $N^{*}$ is an intrinsic property of the signed point cloud and has no free parameters, no regularization, and no train/test split, this quantity avoids the training-related confounds that can make probes optimistic in high-dimensional spaces.

\paragraph{Linear Probes}
To complement the capacity measure, we train two linear classifiers (Hard-Margin SVM and Logistic Regression) on the same activations. The hard-margin SVM also yields a maximum-margin hyperplane whose normal direction we use as a representative encoding direction in \cref{sec:directions}.

\begin{figure*}[h] 
  \begin{center}
    \centerline{\includegraphics[width=\textwidth]{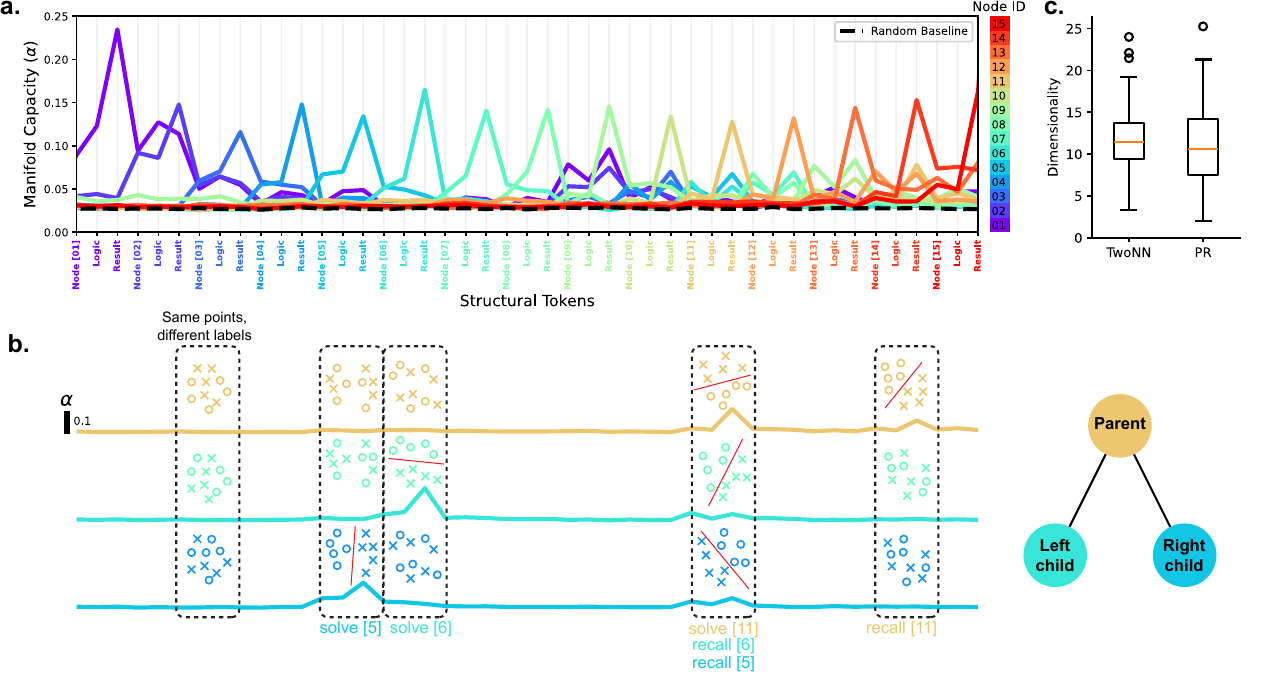}}
    \caption{
    \textbf{Dynamic Modulation of Manifold Geometry during CoT.}
    \textbf{a.} Manifold capacity ($\alpha$) tracked across the Chain-of-Thought sequence (layer 20, tree height $h=4$; height-5 traces in \cref{fig:capacity_trace_SI}). Lines are colored by node ID. Capacity for a specific node peaks sharply at two distinct moments: first when the node is intrinsically computed, and second when it is processed by its parent.
    \textbf{b.} Detailed analysis of Node 11 (yellow) and its children, Node 5 (blue) and Node 6 (turquoise). Insets visualize the latent geometry at key timesteps. High capacity peaks correspond to linearly separable manifolds (clear decision boundaries). Notably, representations become entangled (low capacity) in the interim periods between the initial solution (solve) and the subsequent retrieval (recall). 
    \textbf{c.} Intrinsic dimensionality estimates of the latent embeddings using Two-Nearest Neighbors (TwoNN) and Participation Ratio (PR).}
    \vspace{-0.5cm}
    \label{capacity_trace}
  \end{center}
\end{figure*}

\section{Results}

We analyze the geometry of the residual stream during reasoning on our two compositional tasks. We first validate task performance across models, then characterize the temporal dynamics of Manifold Capacity and Probe Accuracy on the Boolean task, transfer those observations to the natural-language eligibility task, and finally report the spatiotemporal, silenced-CoT, and hyperplane-direction analyses on the Boolean setup.

\subsection{Task Performance}\label{sec:task_perf}
On the depth-5 Boolean Logic task, Ministral 3 8B Reasoning achieves 251/256 all-node trace accuracy (98.0\%) and 254/256 final-answer accuracy (99.2\%) with full CoT generation, compared to 59\% final-answer accuracy when prompted for an immediate answer (cf.\ Fig.~\ref{fig:tree_accuracy}b; the no-CoT prompt is provided in \cref{sec:app_prompts}). This confirms that CoT is functionally necessary for this architecture to solve the task, and ensures that our primary analysis reflects mostly correct reasoning traces. On the full 256-example Boolean replication runs, Qwen2.5-14B-Instruct achieves 248/256 final-answer accuracy (96.9\%) and 242/256 all-node trace accuracy (94.5\%). gpt-oss-20b is less behaviorally stable: it reaches 200/256 final-answer accuracy (78.1\%) and 170/256 all-node trace accuracy (66.4\%), with 192/256 outputs satisfying the strict formatting requirements. We therefore treat gpt-oss-20b as a qualitative geometric replication rather than a high-accuracy behavioral model.

On the eligibility task, Ministral 3 8B Reasoning reaches 256/256 final-answer accuracy and 256/256 all-node accuracy under both the standard prompt and the silenced group-logic prompt. In the unstructured eligibility variant, Ministral reaches 213/256 final-answer accuracy (83.2\%) and 197/256 all-node line-label accuracy (77.0\%). Qwen2.5-7B-Instruct is weaker in this unstructured setting, reaching 166/256 final-answer accuracy (64.8\%) and 111/256 all-node line-label accuracy (43.4\%). Full formatting and perfect-trace counts are reported in \cref{tab:full_run_accuracy}.

\subsection{Dynamic Manifold Management}\label{sec:dynamic_management_boolean}

We analyze Boolean traces from Ministral 3 8B Reasoning at layer 20, a mid-to-late layer where the strongest computational signatures appear in our spatiotemporal analysis (\cref{layer_analysis}); the equivalent dynamics on the eligibility task and the additional models are shown in \cref{sec:eligibility_results,sec:app_boolean_models,sec:app_elig_variants}.

\paragraph{Computation Signal}
We observe that linear separability is not a static property of the representation but follows a distinct pulse-like trajectory. As shown in panels a and b of \cref{capacity_trace}, the capacity computed by grouping examples according to a given node's ground-truth label (the ``node capacity'') spikes when that node is being solved, and then decays back to baseline. This contrasts with the alternative possibility that each node's capacity remains high after the node is solved. In this sense, the node's solution is no longer maintained as a highly separable direction in the residual stream immediately after the computation, creating bandwidth for the subsequent node. Later, when the node is recalled to solve its parent node, the capacity spikes again, but to a lesser degree.

Since the LLM manages the representations such that only a small number of relevant concepts are separable (the two children node answers, and the current node's answer) at a given time, the dimensionality of the representation can be kept small, assuming that each concept takes different directions according to the Linear Representation Hypothesis. This is reflected by our estimations of both the local (intrinsic) and global dimensionalities, measured by the TwoNN \citep{ansuini2019intrinsic} and participation ratios, respectively (\cref{capacity_trace}c). Across the entire CoT, both dimensionalities hover around just 10.

\begin{figure}[t] 
  \begin{center}
    \centerline{\includegraphics[width=\columnwidth]{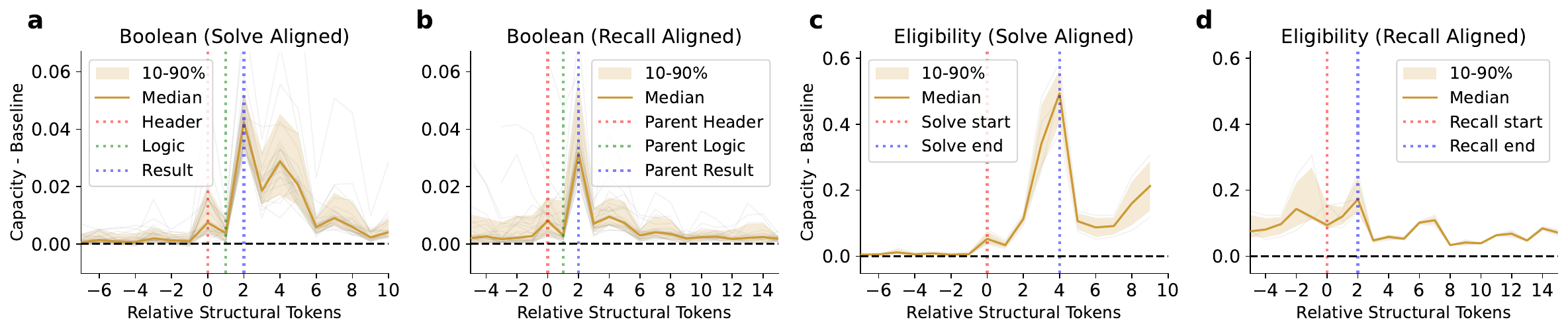}}
    \caption{
    \textbf{Aligned manifold capacity dynamics on both tasks.} Data is aligned to the moment a node is computed (left column: \textbf{a}, \textbf{c}) or when it is recalled by its parent (right column: \textbf{b}, \textbf{d}). (\textbf{a}, \textbf{b}) \emph{Boolean task} (Ministral, layer 20): node-specific capacity shows a sharp, transient peak at the moment of computation and again at recall, dropping quickly to baseline in between. (\textbf{c}, \textbf{d}) \emph{Eligibility task} (Ministral, layer 20): the same qualitative pulse appears on the natural-language task. Vertical dashed lines mark the positions of structural anchors within each phase. The corresponding capacity vs.\ trained-probe accuracy comparison on the Boolean task is shown in \cref{sec:app_probe_divergence}.
    }
    \label{fig:aligned}
  \end{center}
\end{figure}

To summarize the canonical capacity trace around solve and recall phases, we align the Boolean traces at the solve-phase structural anchors (\cref{fig:aligned}a) and at the recall-phase structural anchors (\cref{fig:aligned}b). Each phase contains three structural anchors: ``header'', ``logic'', and ``result''. The capacity for a target node peaks at the anchors immediately \textit{preceding} the generation of the textual answer. The increase in separability begins as soon as the model generates the tokens corresponding to the node's dependencies. This indicates that the linear separability of the concept emerges in the residual stream prior to the explicit token decoding. The same qualitative pulse pattern is observed on the Boolean task with Qwen2.5-14B-Instruct and gpt-oss-20b; the per-model traces are shown in \cref{sec:app_boolean_models}.

\paragraph{Separability Decay and Divergence with Linear Probes}
Immediately after a node's computation is complete, its Manifold Capacity decays rapidly (\cref{fig:aligned}a,b). This marks a divergence from linear probes trained on the same features: while probe accuracy remains high ($>90\%$) for the tokens corresponding to the next two nodes being computed, Manifold Capacity drops near the random baseline. This divergence highlights a difference between information that is recoverable by a trained linear readout and information that is intrinsically organized as a high-capacity, linearly separable manifold in the current geometry. The aligned probe-vs-capacity comparison is reported in \cref{sec:app_probe_divergence}; the probe signal is noisier and weaker than capacity and is most clearly visible only when the traces are aligned.

\subsection{Generalization to a Natural-Language Task}\label{sec:eligibility_results}

We find that Dynamic Manifold Management extends to a more natural compositional reasoning task. The eligibility task preserves the same known tree structure needed for temporal alignment, while replacing symbolic leaf values with prose-based evidence extraction and threshold comparison: each leaf computation requires reading the applicant paragraph, locating the relevant attribute, and comparing it to a numerical threshold.

The same pulse pattern appears in this setting (panels c and d of \cref{fig:aligned}, Ministral on the standard eligibility prompt). At the solve-aligned anchors (\cref{fig:aligned}c), the capacity for a given criterion or group node rises as that node is being computed and then decays. At the recall-aligned anchors (\cref{fig:aligned}d), the children of the current parent node show a renewed capacity rise during parent computation, mirroring the Boolean recall signal. The corresponding full traces are shown in \cref{sec:app_elig_variants}.

We also evaluate Ministral on an \emph{unstructured} version of the eligibility task in which the model writes free natural-language reasoning lines, with only lightweight ID prefixes (\texttt{C1:}, \dots, \texttt{Final:}) marking each step (full prompt and example response in \cref{sec:app_prompts,sec:app_eligibility_example}). For this variant we sample within-line word-end anchors as temporal alignment points instead of fixed Markdown markers. The pulse and recall pattern is preserved for Ministral (\cref{fig:elig_unstructured}; 83.2\% final-answer accuracy and 77.0\% all-node line-label accuracy), and the same qualitative behavior is observed when the unstructured variant is run on Qwen2.5-7B-Instruct (\cref{fig:elig_qwen7b_unstructured}), despite lower task performance on that model (64.8\% final-answer accuracy and 43.4\% all-node line-label accuracy). Thus, the geometric signature is not tied to the structured template or to one specific model, although the Qwen2.5-7B-Instruct result should be interpreted as a lower-accuracy replication.

\subsection{Spatiotemporal Localization}
We extended our analysis to the full Layer $\times$ Time grid to locate the physical origin of these geometric pulses. Because the residual stream is recomputed at every token, this lets us see, for each piece of information, which layers expand or compress it at each step.

\begin{figure}[t]
  \def\mysize{\columnwidth}
  \begin{center}
    \includegraphics[width=0.5\mysize]{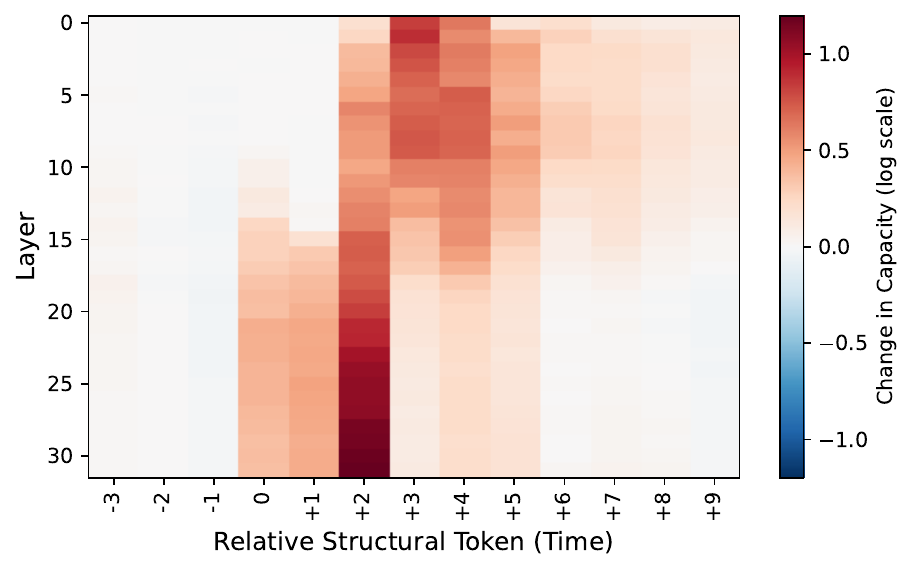}
    \label{one_capacity_trace}
    \caption{
    \textbf{Spatiotemporal heatmap of manifold capacity change.} The x-axis represents token positions relative to the computation step ($t=0$). (\emph{During Computation, $t=0,1$}) A capacity increase initiates in the middle layers at $t=0$, culminating in a sharp peak within the deep layers at the token preceding the result ($t=+2$). (\emph{Post-Computation}) For $t > +2$, the early-to-mid layers retain elevated capacity as they retain the immediate context; however, this capacity decreases in the middle layers as the model reorganizes its representation to shift focus to the next node.
    }
    \label{layer_analysis}
  \end{center}
\end{figure}

\Cref{layer_analysis} presents the capacity change averaged over all nodes along the reasoning trace, aligned with respect to the moment where the node is solved. Two distinct phases emerge: during the \emph{reasoning phase}, the increase in separability originates in the middle layers (13--15) at the header and logic anchors and then propagates through all layers at the result anchor; during the subsequent \emph{reorganization phase}, the middle layers rapidly suppress capacity for the previous node while the early layers retain it as part of the immediate context, leaving the information no longer readily accessible in the deeper layers (16--32). The linear-probe heatmap (\cref{fig:svm_heatmap}) shows a similar reasoning phase but a much less pronounced suppression: probe accuracy stays above 90\% in later layers for up to two nodes past the computation.

\subsection{Implicit Computation under Silenced CoT}\label{sec:silenced}

Does the capacity of a node rise during recall simply because the model is about to write that node's truth value (or even its name) into the output, or because of a computational need? We test this on both tasks by silencing the part of the textual output that mentions the child labels during parent computation: on the Boolean task we suppress the explicit truth values (and, in a stronger variant, the entire \texttt{Logic} expression) of the children at the parent step (see \cref{sec:silent_cot1}), and on the eligibility task we mask the \texttt{Logic} lines of the group nodes ($G_1$, $G_2$, \texttt{Final}) so the parent step cannot explicitly mention the child labels (silenced eligibility figures are shown in \cref{sec:app_elig_variants}). In both tasks, the child-node capacities still spike during the recall phase under silencing, arguing that the recall-time geometric pulse is driven by the internal computation rather than by the model preparing to literally emit the child token.

\subsection{Directions of Separation}\label{sec:directions}

We saw in \cref{capacity_trace}c that the dimensionality is kept low during CoT, and only a few concepts are separable, minimizing the bandwidth of the residual stream. To understand which variables occupy this bandwidth during computation, we investigate the normal directions of the max-margin separating hyperplanes.

\cref{capacity_trace} shows that when a node is being solved, the manifold is separable by the answers of both the left and right child nodes. Interestingly, we find two strongly conserved directions for separating the left and right child-node answers, and these directions are slightly anti-aligned, requiring at least two dimensions (\cref{fig:inter_node}-\textbf{right}). The conservation of the left- and right-child directions suggests that the LLM solves the problem compositionally, applying the same local rule across subproblems. Since the manifold also needs to be separated with respect to the parent node, the minimum number of dimensions required for the computation is three. In addition, we find that the direction separating a node's answer while it is being solved is not conserved across nodes (\cref{fig:inter_node}-\textbf{left}). Moreover, the directions used when a node is solved and recalled are also not conserved (\cref{fig:within_node}). See \cref{sec:supp_direction} for more details.

\section{Discussion}

\paragraph{Geometric Signatures of Active Computation}
Motivated by the open question of whether CoT is a post-hoc rationalization or reflects active internal computation \citep{turpin2023faithful}, our results favor the active-computation interpretation. We observe a \emph{pre-computation signal} where the manifold capacity for a specific logical node peaks immediately prior to the generation of the answer tokens. This precise temporal synchronization is consistent with the model actively modulating the geometry of its residual stream to render specific information linearly accessible exactly when the computation requires it. This pulsing behavior acts as a geometric signature of the reasoning process, distinguishing functional computation from inert representation.

\paragraph{Retention vs. Readiness of Information}
A key finding of our study is the point where Manifold Capacity and standard Linear Probes diverge: probes maintain high accuracy long after a computational step is finished, while Manifold Capacity decays rapidly. We propose that these two metrics capture different states of information. High probe accuracy indicates \emph{retention}: the information remains encoded in the high-dimensional latent space. High manifold capacity indicates \emph{readiness}: the information is explicitly structured for linear readout, enabling downstream computation. Combined with the direction analysis, this rapid decay of capacity suggests a mechanism of subspace rotation: once a logical step is over, the model rotates the representation out of the primary linear subspace used for computation. This effectively clears the active workspace, preventing past variables from interfering with the current logical operation. Past information does not need to be fully suppressed and can still be recovered by a linear probe, but it is effectively archived and not ready for computation.

\paragraph{Dynamic Manifold Management}
We interpret these combined dynamics, the localized pulses, the subspace rotation, the layer specialization revealed by the spatiotemporal analysis, and the strong correlation between cross-token attention and source-node capacity (\cref{sec:app_attention}), as a unified mechanism of \emph{Dynamic Manifold Management}. For reasoning tasks with high compositionality, the fixed dimensionality of the residual stream becomes a central constraint. A static representation of a complex reasoning tree would require maintaining linearly separable directions for every intermediate variable simultaneously, quickly exhausting the available spatial bandwidth. Instead, CoT allows the model to distribute computational load across time. It optimizes the geometry of its residual stream by expanding relevant concepts into separable manifolds \textit{only} during the moment of computation, and compressing them immediately afterward to save space and attend to the next computation without interference. This supports the view of the residual stream as a global workspace: the model uses the generated text to offload information, retrieving and processing it in the middle layers only when necessary.

\section*{Limitations}
Our deepest mechanistic analyses (spatiotemporal localization, hyperplane directions, attention) rely on the Boolean tree task, which is symbolic and parametric and gives us the structural control needed for those analyses. The eligibility task and the additional models extend the central pulse and recall observation beyond that controlled setting, but they are still small-to-medium open-weight models on tasks that have a fixed compositional tree. Whether the same dynamics persist on open-ended reasoning tasks such as competition math or program synthesis, and on larger models with wider residual streams (e.g., 70B+ parameters), remains to be tested. Our reliance on Manifold Capacity also assumes the model primarily uses linear separability for internal steps. While the Linear Representation Hypothesis is well supported empirically, LLMs may also rely on non-linear computations that our current metrics do not fully capture; the divergence we report between capacity and probe accuracy is informative regardless, but it does not rule out additional non-linear structure.

\begin{ack}
We thank Abdulkadir Canatar for early discussions, Yonatan Belinkov and Kyunghyun Cho for insightful feedback, and Manu Srinath Halvagal and Artem Kirsanov for helpful comments. This work is supported by the Samsung Advanced Institute of Technology (under the project ``Next Generation Deep Learning: From Pattern Recognition to AI''). S.C. is also supported by the Klingenstein-Simons Award, and a Sloan Research Fellowship.
\end{ack}

\bibliography{biblio}
\bibliographystyle{plainnat}

\newpage
\appendix
\onecolumn
\setcounter{figure}{0} 
\renewcommand{\thefigure}{S.\arabic{figure}} 
\setcounter{table}{0} 
\renewcommand{\thetable}{S.\arabic{table}}

\section{Prompts}\label{sec:app_prompts}

This appendix collects the system prompts used in the experiments. Each prompt was used verbatim as a system message; the per-item user message is the task instance (a Boolean expression or an applicant description with its criteria).

\subsection{Boolean Task: Standard CoT Prompt (Ministral)}

\begin{PromptVerbatim}
"You are a precise boolean logic solver. You will be given a boolean expression "
"where specific operations are labeled with IDs like [01], [02], etc.\n\n"
"Your task is to solve the tree step-by-step in strictly increasing order of these IDs.\n"
"Follow the exact format shown in the example below.\n\n"
"### EXAMPLE INPUT:\n"
"Is the following boolean expression True or False?\n"
"[03]: ([01]: (True or False) and [02]: (False xor True))\n\n"
"### EXAMPLE OUTPUT:\n"
"### Problem Statement\n"
"1. **Expression**: `[03]: ([01]: (True or False) and [02]: (False xor True))`\n"
"2. **Node IDs**: [01], [02], [03]\n\n"
"### Solve\n"
"**Node [01]**\n"
"* Logic: `True or False`\n"
"* Result: `True`\n\n"
"**Node [02]**\n"
"* Logic: `False xor True`\n"
"* Result: `True`\n\n"
"**Node [03]**\n"
"* Logic: `[01] and [02]` -> `True and True`\n"
"* Result: `True`\n\n"
"### Summary\n"
#"* [01]: True\n"
#"* [02]: True\n"
#"* [03]: True\n"
"**Final Answer: True**\n\n"
"### YOUR TURN:\n"
\end{PromptVerbatim}

\subsection{Boolean Task: Strict-Format CoT Prompt (Qwen2.5-14B-Instruct, gpt-oss-20b)}

For the additional Boolean models, the standard prompt was extended with stricter formatting rules to keep the output usable for activation-anchor extraction.

\begin{PromptVerbatim}
You are a precise boolean logic solver.

You will be given a boolean expression where operations are labeled with IDs like [01], [02], etc.

Follow these rules exactly:
1. Solve the nodes strictly in increasing numeric order.
2. Copy every node ID exactly as it appears in the expression, including brackets and any leading zeros.
3. Do not skip any node.
4. Start your answer immediately with `### Problem Statement`.
5. Use the exact headings, punctuation, and bullet structure shown in the example, including `2. **Node IDs**:` with the colon.
6. End with `**Final Answer: True**` or `**Final Answer: False**`.
7. Do not add any extra commentary before or after the required format.

### EXAMPLE INPUT:
Is the following boolean expression True or False?
[03]: ([01]: (True or False) and [02]: (False xor True))

### EXAMPLE OUTPUT:
### Problem Statement
1. **Expression**: `[03]: ([01]: (True or False) and [02]: (False xor True))`
2. **Node IDs**: [01], [02], [03]

### Solve
**Node [01]**
* Logic: `True or False`
* Result: `True`

**Node [02]**
* Logic: `False xor True`
* Result: `True`

**Node [03]**
* Logic: `[01] and [02]` -> `True and True`
* Result: `True`

**Final Answer: True**

### YOUR TURN:
\end{PromptVerbatim}

\subsection{Boolean Task: Masked-Logic Prompt (silenced variant)}

This prompt defines the Boolean silenced variant used in \cref{sec:silenced}. Each \texttt{* Logic:} line is masked with exactly fifteen asterisks inside backticks; the model still has to produce the correct \texttt{* Result:} for each node.

\begin{PromptVerbatim}
You are a precise boolean logic solver. You will be given a boolean expression 
where specific operations are labeled with IDs like [01], [02], etc.

Your task is to solve the tree step-by-step in strictly increasing order of these IDs.
Follow the exact format shown in the example below.

CRITICAL INSTRUCTION: In the 'Logic' step, do not write the boolean values. Instead, mask the computation using exactly 15 asterisks inside backticks, like `***************`.

### EXAMPLE INPUT:
Is the following boolean expression True or False?
[03]: ([01]: (True or False) and [02]: (False xor True))

### EXAMPLE OUTPUT:
### Problem Statement
1. **Expression**: `[03]: ([01]: (True or False) and [02]: (False xor True))`
2. **Node IDs**: [01], [02], [03]

### Solve
**Node [01]**
* Logic: `***************`
* Result: `True`

**Node [02]**
* Logic: `***************`
* Result: `True`

**Node [03]**
* Logic: `***************`
* Result: `True`

### Summary
* [01]: True
* [02]: True
* [03]: True
**Final Answer: True**

### YOUR TURN:
\end{PromptVerbatim}

\subsection{Boolean Task: No-CoT Prompt}

This is the prompt used to elicit the no-CoT baseline accuracy reported in \cref{sec:task_perf}.

\begin{PromptVerbatim}
"You are a precise boolean logic calculator.\n"
"You will be given a boolean expression where specific operations are labeled with IDs.\n"
"Your task is to evaluate the expression and output ONLY the final True/False result.\n\n"
"**RULES:**\n"
"1. DO NOT think step-by-step.\n"
"2. DO NOT explain your reasoning.\n"
"3. DO NOT list node IDs or intermediate logic.\n"
"4. Respond immediately with the final answer.\n\n"
"Follow the exact format shown in the example below.\n\n"
"### EXAMPLE INPUT:\n"
"Is the following boolean expression True or False?\n"
"[03]: ([01]: (True or False) and [02]: (False xor True))\n\n"
"### EXAMPLE OUTPUT:\n"
"**Final Answer: True**\n\n"
"### YOUR TURN:\n"
\end{PromptVerbatim}

\subsection{Eligibility Task: Standard Prompt}

\begin{PromptVerbatim}
You are a precise eligibility evaluator. You will be given an applicant's profile as a short text description, along with eligibility criteria organized in a tree structure with IDs like C1, C2, G1, G2, etc.

Your task is to evaluate each criterion and group in strictly increasing order of their IDs (C1, C2, G1, C3, C4, G2, Final). Follow the exact format shown in the example below.

### EXAMPLE INPUT:
Evaluate whether the following applicant meets the eligibility criteria.

**Applicant:**
Maria Chen is a 29-year-old graphic designer from Denver, Colorado. She earns $74,000 per year and has been at her current firm for 4 years. Her credit score is 710, and she has $18,500 in liquid savings. She has no dependents and holds a bachelor's degree.

**Criteria:**
- C1 (Age): age ≥ 27
- C2 (Income): income ≥ $70,000
- G1 (Financial Stability): C1 AND C2
- C3 (Credit Score): credit score ≥ 720
- C4 (Savings): savings ≥ $15,000
- G2 (Credit Profile): C3 OR C4
- Final (Eligible): G1 AND G2

### EXAMPLE OUTPUT:
### Evaluation

**Criterion C1 (Age)**
* Extract: `29`
* Threshold: `≥ 27`
* Compare: `29 ≥ 27`
* Result: `Met`

**Criterion C2 (Income)**
* Extract: `$74,000`
* Threshold: `≥ $70,000`
* Compare: `$74,000 ≥ $70,000`
* Result: `Met`

**Group G1 (Financial Stability)**
* Logic: `C1 AND C2` -> `Met AND Met`
* Result: `Met`

**Criterion C3 (Credit Score)**
* Extract: `710`
* Threshold: `≥ 720`
* Compare: `710 ≥ 720`
* Result: `Not Met`

**Criterion C4 (Savings)**
* Extract: `$18,500`
* Threshold: `≥ $15,000`
* Compare: `$18,500 ≥ $15,000`
* Result: `Met`

**Group G2 (Credit Profile)**
* Logic: `C3 OR C4` -> `Not Met OR Met`
* Result: `Met`

**Final (Eligible)**
* Logic: `G1 AND G2` -> `Met AND Met`
* Result: `Met`

### Summary
* C1: Met
* C2: Met
* G1: Met
* C3: Not Met
* C4: Met
* G2: Met
* Final: Met

### YOUR TURN:
\end{PromptVerbatim}

\subsection{Eligibility Task: Silenced Group-Logic Prompt}

This prompt defines the silenced eligibility variant used in \cref{sec:silenced}. The criterion phases are unchanged; only the \texttt{* Logic:} lines for the group nodes ($G_1$, $G_2$, \texttt{Final}) are masked, so the parent step cannot explicitly mention the child labels or operators.

\begin{PromptVerbatim}
You are a precise eligibility evaluator. You will be given an applicant's profile as a short text description, along with eligibility criteria organized in a tree structure with IDs like C1, C2, G1, G2, etc.

Your task is to evaluate each criterion and group in strictly increasing order of their IDs (C1, C2, G1, C3, C4, G2, Final). Follow the exact format shown in the example below.

CRITICAL INSTRUCTION: Do not mask the criterion phases. For the group and final `* Logic:` lines only, do not reveal the child labels, operators, or intermediate boolean-style composition. Instead, write exactly six asterisks inside backticks: `******`.

### EXAMPLE INPUT:
Evaluate whether the following applicant meets the eligibility criteria.

**Applicant:**
Maria Chen is a 29-year-old graphic designer from Denver, Colorado. She earns $74,000 per year and has been at her current firm for 4 years. Her credit score is 710, and she has $18,500 in liquid savings. She has no dependents and holds a bachelor's degree.

**Criteria:**
- C1 (Age): age ≥ 27
- C2 (Income): income ≥ $70,000
- G1 (Financial Stability): C1 AND C2
- C3 (Credit Score): credit score ≥ 720
- C4 (Savings): savings ≥ $15,000
- G2 (Credit Profile): C3 OR C4
- Final (Eligible): G1 AND G2

### EXAMPLE OUTPUT:
### Evaluation

**Criterion C1 (Age)**
* Extract: `29`
* Threshold: `≥ 27`
* Compare: `29 ≥ 27`
* Result: `Met`

**Criterion C2 (Income)**
* Extract: `$74,000`
* Threshold: `≥ $70,000`
* Compare: `$74,000 ≥ $70,000`
* Result: `Met`

**Group G1 (Financial Stability)**
* Logic: `******`
* Result: `Met`

**Criterion C3 (Credit Score)**
* Extract: `710`
* Threshold: `≥ 720`
* Compare: `710 ≥ 720`
* Result: `Not Met`

**Criterion C4 (Savings)**
* Extract: `$18,500`
* Threshold: `≥ $15,000`
* Compare: `$18,500 ≥ $15,000`
* Result: `Met`

**Group G2 (Credit Profile)**
* Logic: `******`
* Result: `Met`

**Final (Eligible)**
* Logic: `******`
* Result: `Met`

### Summary
* C1: Met
* C2: Met
* G1: Met
* C3: Not Met
* C4: Met
* G2: Met
* Final: Met

### YOUR TURN:
\end{PromptVerbatim}

\subsection{Eligibility Task: Unstructured Prompt}

This prompt defines the unstructured variant used in \cref{sec:eligibility_results}. The model is asked to write free natural-language reasoning lines, with only lightweight ID prefixes (\texttt{C1:}, \dots, \texttt{Final:}) marking each step. Within-line word-end positions are used as temporal alignment anchors.

\begin{PromptVerbatim}
You are a careful eligibility evaluator.

You will be given an applicant description and a set of named eligibility items: C1, C2, G1, C3, C4, G2, and Final.

Reason in natural language and decide whether the applicant is eligible.

Important instructions:
- Evaluate all of the named items before finishing.
- For each item, write exactly one line that begins with its ID followed by a colon.
- Use these seven line starters exactly once each:
  C1:
  C2:
  G1:
  C3:
  C4:
  G2:
  Final:
- After the ID and colon, you may write freely in natural language.
- Use the exact numbers and thresholds given in the applicant description and criteria. Do not change, round, or invent any values.
- After finishing the seven reasoning lines, write exactly one line in one of these two forms:
  Final answer: Eligible
  Final answer: Not Eligible
- Then write exactly one line in this form:
  Violated criteria: <comma-separated list of all named items that were not met>
- If no named item was violated, write:
  Violated criteria: none
- The `Violated criteria` line must use the same seven named items only: C1, C2, G1, C3, C4, G2, Final.
- If the applicant is not eligible, include `Final` in the violated list.

### EXAMPLE INPUT:
Evaluate whether the following applicant meets the eligibility criteria.

**Applicant:**
Maria Chen is a 29-year-old graphic designer from Denver, Colorado. She earns $74,000 per year and has been at her current firm for 4 years. Her credit score is 710, and she has $18,500 in liquid savings. She has no dependents and holds a bachelor's degree.

**Criteria:**
- C1 (Age): age ≥ 27
- C2 (Income): income ≥ $70,000
- G1 (Financial Stability): C1 AND C2
- C3 (Credit Score): credit score ≥ 720
- C4 (Savings): savings ≥ $15,000
- G2 (Credit Profile): C3 OR C4
- Final (Eligible): G1 AND G2

### EXAMPLE OUTPUT:
C1: Maria Chen is 29 years old, which is at least 27, so this item is met.
C2: She earns $74,000 per year, which is at least $70,000, so this item is met.
G1: The required condition for this item is satisfied, so this item is met.
C3: Her credit score is 710, which is below 720, so this item is not met.
C4: She has $18,500 in liquid savings, which is at least $15,000, so this item is met.
G2: The required condition for this item is satisfied, so this item is met.
Final: The final eligibility condition is satisfied, so the applicant is eligible.

Final answer: Eligible
Violated criteria: C3

### YOUR TURN:
\end{PromptVerbatim}

\section{Eligibility Task: Example Input and Response}\label{sec:app_eligibility_example}

\subsection{Example Input}

\begin{PromptVerbatim}
Evaluate whether the following applicant meets the eligibility criteria.

**Applicant:**
Brian Hawkins, age 41, is a financial analyst based in Nashville,
Tennessee. He has been with his current position for 8 years. He holds a
professional degree. He earns $63,000 per year. He maintains $29,500 in
readily accessible savings. He has a credit score of 730.

**Criteria:**
- C1 (Age): age >= 40
- C2 (Income): income >= $60,000
- G1 (Financial Stability): C1 AND C2
- C3 (Credit Score): credit score >= 720
- C4 (Savings): savings >= $25,000
- G2 (Credit Profile): C3 AND C4
- Final (Eligible): G1 OR G2
\end{PromptVerbatim}

\subsection{Example Response (Ministral, Unstructured Prompt)}

\begin{PromptVerbatim}
C1: Brian Hawkins is 41 years old, which meets the requirement of age >= 40.
C2: He earns $63,000 per year, which is at least $60,000, so this item is
    met.
G1: Both C1 and C2 are met, so this item is met.
C3: His credit score is 730, which is at least 720, so this item is met.
C4: He has $29,500 in savings, which is at least $25,000, so this item
    is met.
G2: Both C3 and C4 are met, so this item is met.
Final: Since both G1 and G2 are met, the applicant is eligible.

Final answer: Eligible
Violated criteria: none
\end{PromptVerbatim}

\section{Per-Model Task Performance}\label{sec:app_accuracy}

\Cref{tab:full_run_accuracy} reports the full 256-example runs used in the main analyses and appendix replications. For the unstructured eligibility setting, ``all nodes'' refers to the seven ID-prefixed reasoning lines ($C_1, C_2, G_1, C_3, C_4, G_2, \texttt{Final}$), and ``perfect'' additionally requires the explicit final-answer and violated-criteria lines to match the ground truth.

\begin{table}[ht]
\centering
\small
\setlength{\tabcolsep}{3pt}
\begin{tabular}{lllcccc}
\toprule
Task & Model & Prompt & Formatting & All nodes & Final & Perfect \\
\midrule
Boolean & Qwen2.5-14B-Instruct & strict-format & 256/256 & 242/256 & 248/256 & 242/256 \\
Boolean & gpt-oss-20b & strict-format & 192/256 & 170/256 & 200/256 & 153/256 \\
Boolean & Ministral 3 8B Reasoning & standard CoT & 256/256 & 251/256 & 254/256 & 251/256 \\
Eligibility & Ministral 3 8B Reasoning & standard & 256/256 & 256/256 & 256/256 & 256/256 \\
Eligibility & Ministral 3 8B Reasoning & silenced group-logic & 256/256 & 256/256 & 256/256 & 256/256 \\
Eligibility & Ministral 3 8B Reasoning & unstructured & 256/256 & 197/256 & 213/256 & 171/256 \\
Eligibility & Qwen2.5-7B-Instruct & unstructured & 255/256 & 111/256 & 166/256 & 54/256 \\
\bottomrule
\end{tabular}
\caption{Full-run behavioral accuracy ($n = 256$ examples per row) for the runs used in the reported analyses. ``All nodes'' means every reported node, criterion, or group label is correct. ``Perfect'' additionally requires strict formatting for the Boolean task, and for unstructured eligibility also requires the final-answer and violated-criteria lines to be correct.}
\label{tab:full_run_accuracy}
\end{table}

\section{Boolean Task: Replication Across Models}\label{sec:app_boolean_models}

This section reports the Boolean dynamics on the two additional models discussed in \cref{sec:dynamic_management_boolean}. The same qualitative pulse and recall pattern that we observe on Ministral (\cref{capacity_trace} and panels a and b of \cref{fig:aligned}) is also present here, on models with different parameter counts and different training pipelines.

\begin{figure}[ht]
  \centering
  \includegraphics[width=0.48\textwidth]{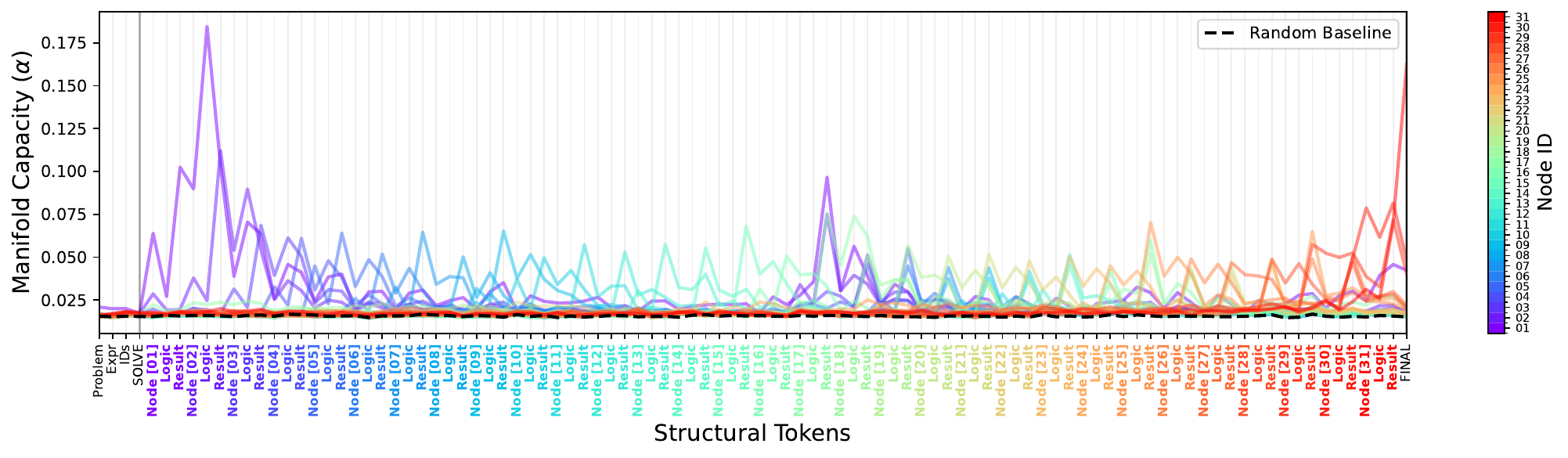}\hfill
  \includegraphics[width=0.48\textwidth]{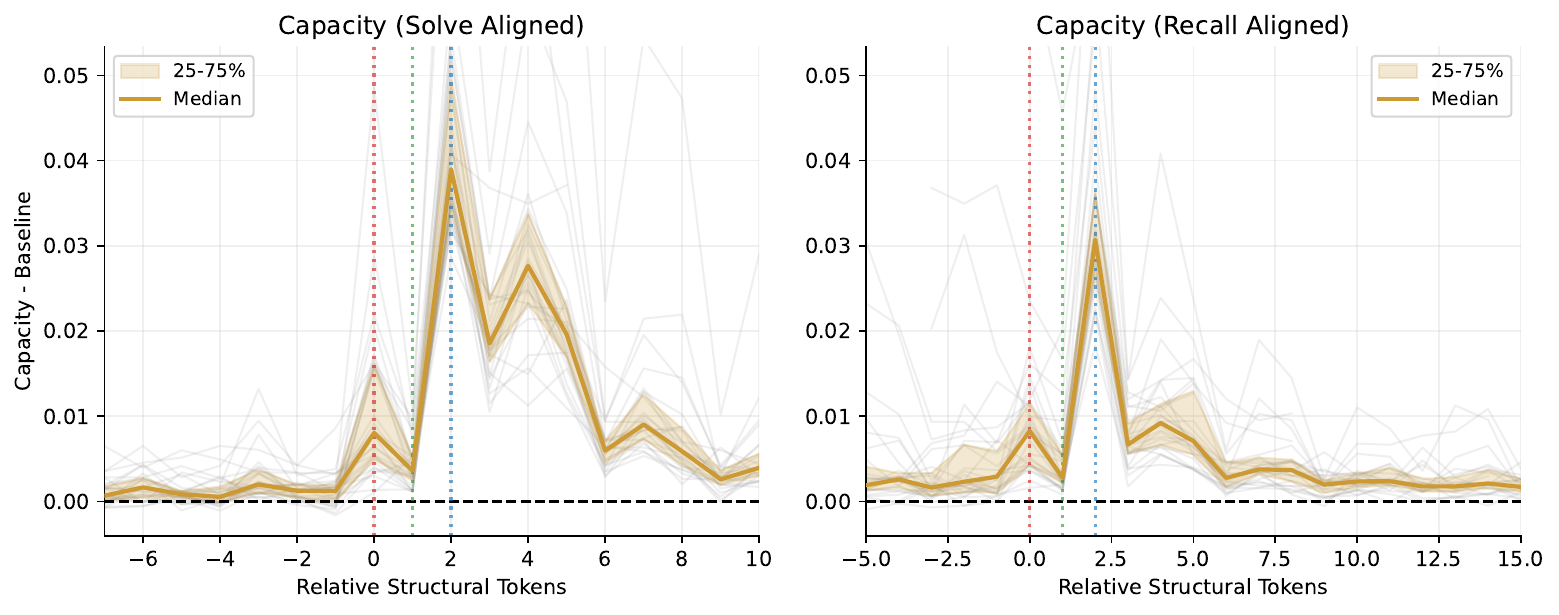}
  \caption{\textbf{Boolean task on Qwen2.5-14B-Instruct (layer 26).} \emph{Left}: full capacity traces over the CoT, colored by node ID. \emph{Right}: solve- and recall-aligned capacity. The pulse and recall pattern is preserved.}
  \label{fig:qwen14b_boolean}
\end{figure}

\begin{figure}[ht]
  \centering
  \includegraphics[width=0.48\textwidth]{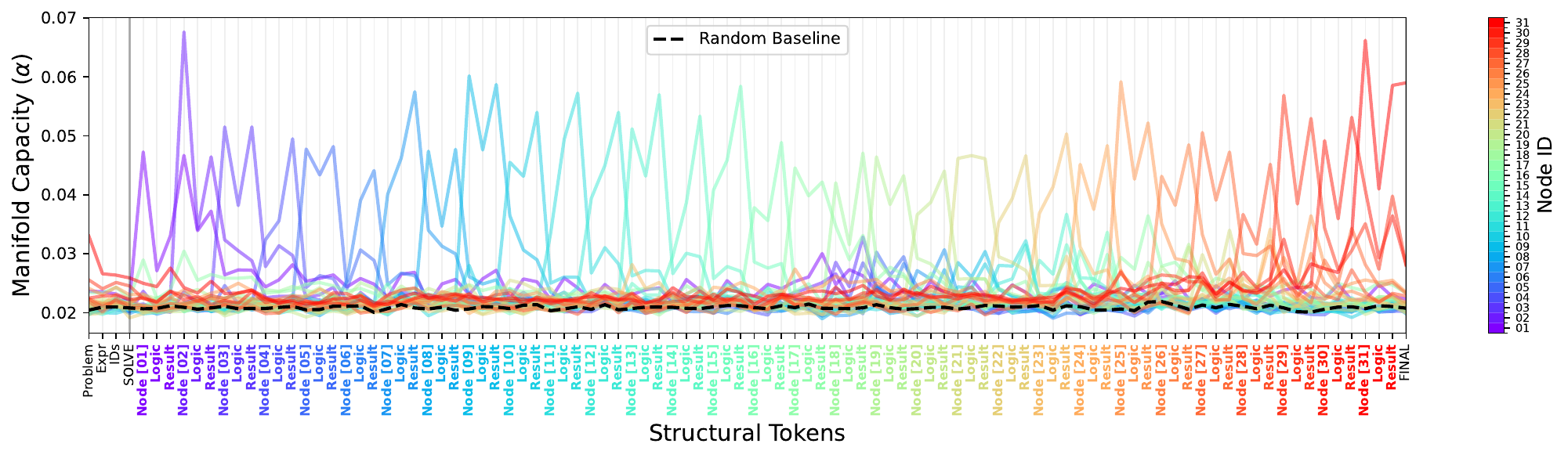}\hfill
  \includegraphics[width=0.48\textwidth]{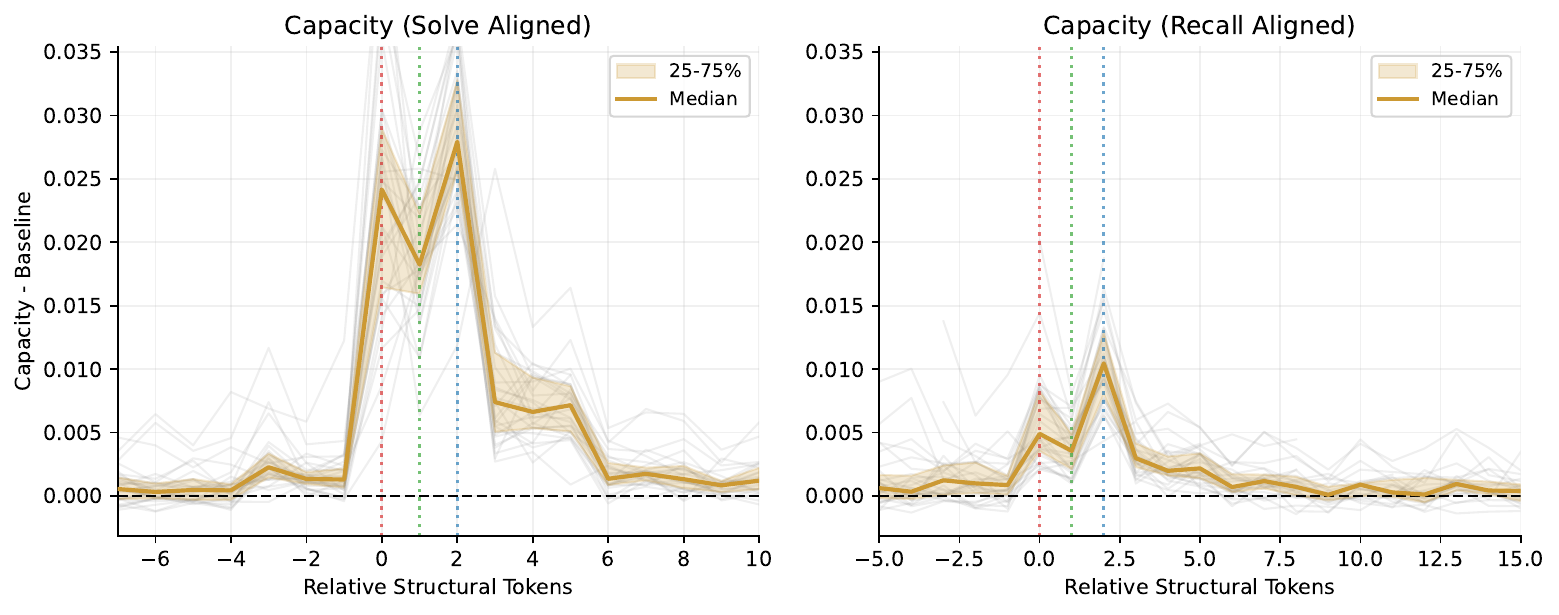}
  \caption{\textbf{Boolean task on gpt-oss-20b (layer 13).} Same conventions as \cref{fig:qwen14b_boolean}.}
  \label{fig:gptoss_boolean}
\end{figure}

\section{Eligibility Task: Variants and Per-Model Replication}\label{sec:app_elig_variants}

This section reports the eligibility-task variants discussed in \cref{sec:eligibility_results,sec:silenced}. Panels c and d of \cref{fig:aligned} show the standard-prompt aligned traces; the panels below show the corresponding full-trace plot for the standard prompt, the silenced group-logic variant (full and aligned), and the unstructured variant (full and aligned). We also evaluate the unstructured variant on Qwen2.5-7B-Instruct.

\begin{figure}[ht]
  \centering
  \includegraphics[width=0.7\textwidth]{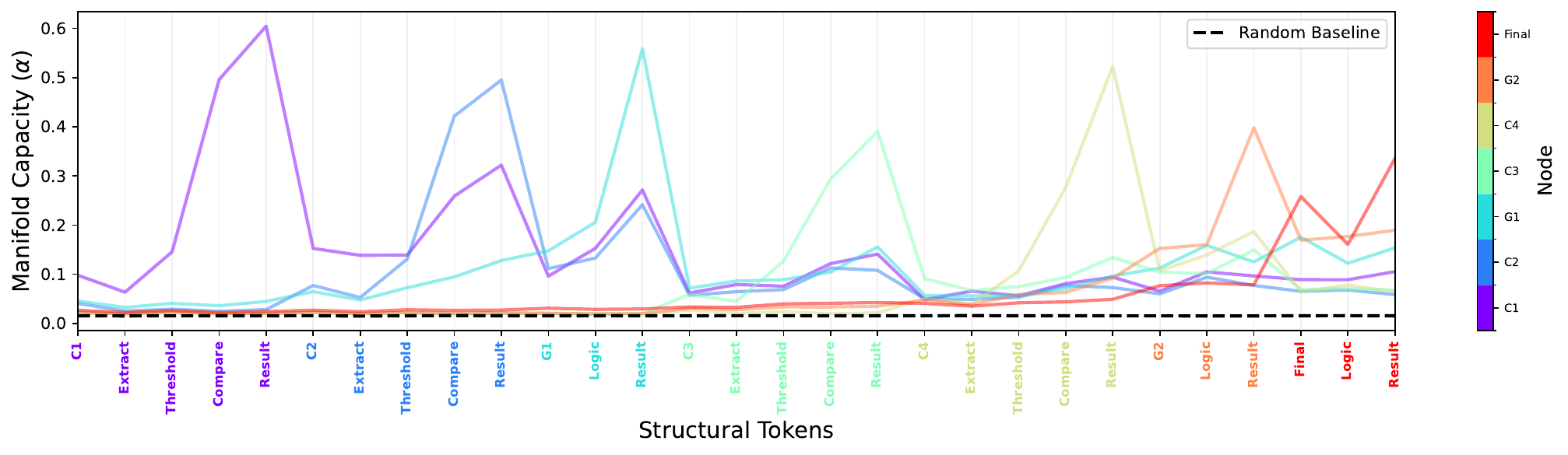}
  \caption{\textbf{Eligibility task, standard prompt, Ministral (layer 20): full traces.} Capacity for each node ID across the full reasoning trace. The aligned solve and recall views are shown in panels c and d of \cref{fig:aligned}.}
  \label{fig:elig_standard_full}
\end{figure}

\begin{figure}[ht]
  \centering
  \includegraphics[width=0.48\textwidth]{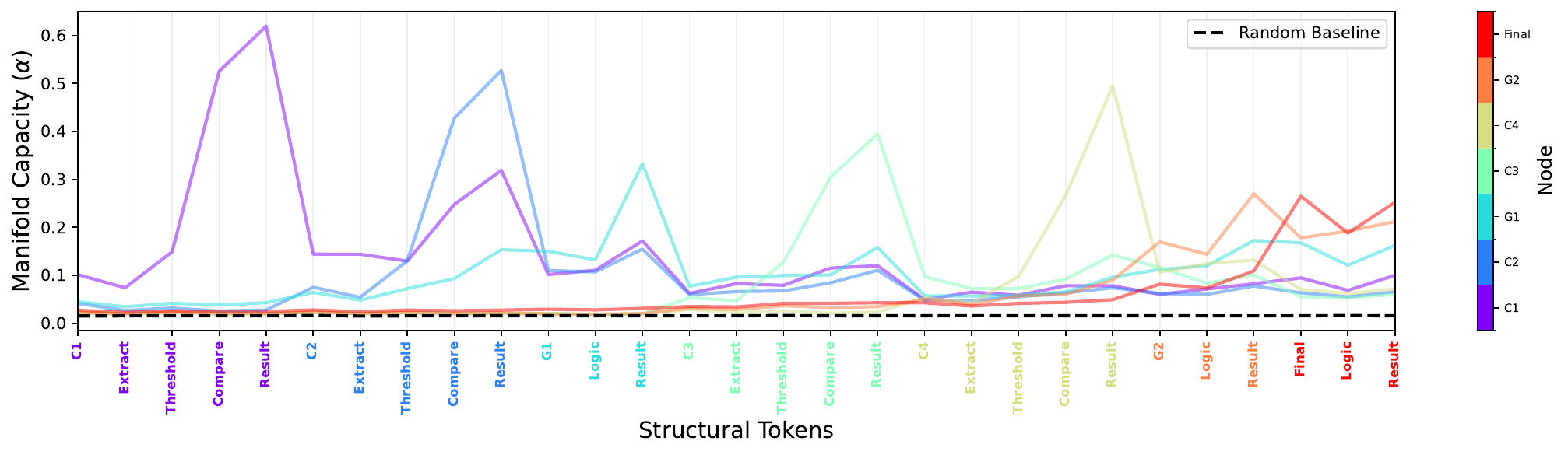}\hfill
  \includegraphics[width=0.48\textwidth]{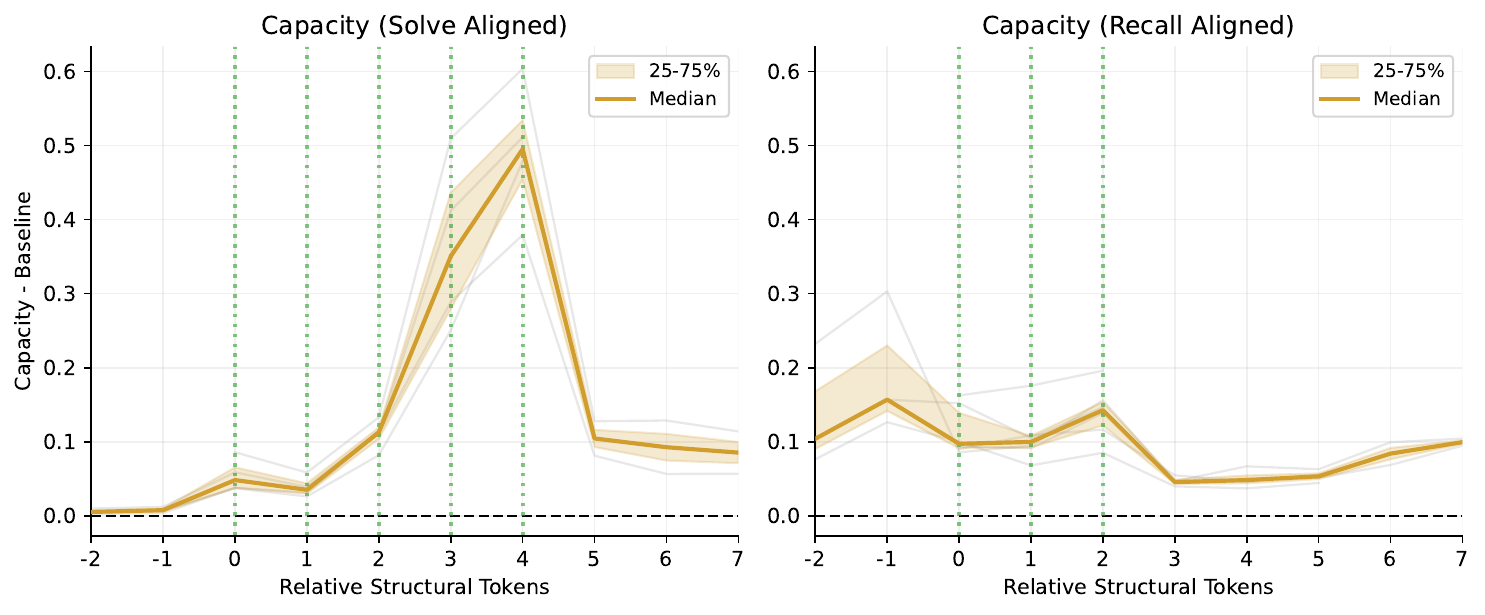}
  \caption{\textbf{Eligibility task, silenced group-logic prompt, Ministral (layer 20).} The \texttt{* Logic:} lines for $G_1$, $G_2$, and \texttt{Final} are masked, so the parent step cannot explicitly mention the child labels. The recall-time capacity rise of the children persists, supporting the conclusion in \cref{sec:silenced}.}
  \label{fig:elig_silenced}
\end{figure}

\begin{figure}[ht]
  \centering
  \includegraphics[width=0.48\textwidth]{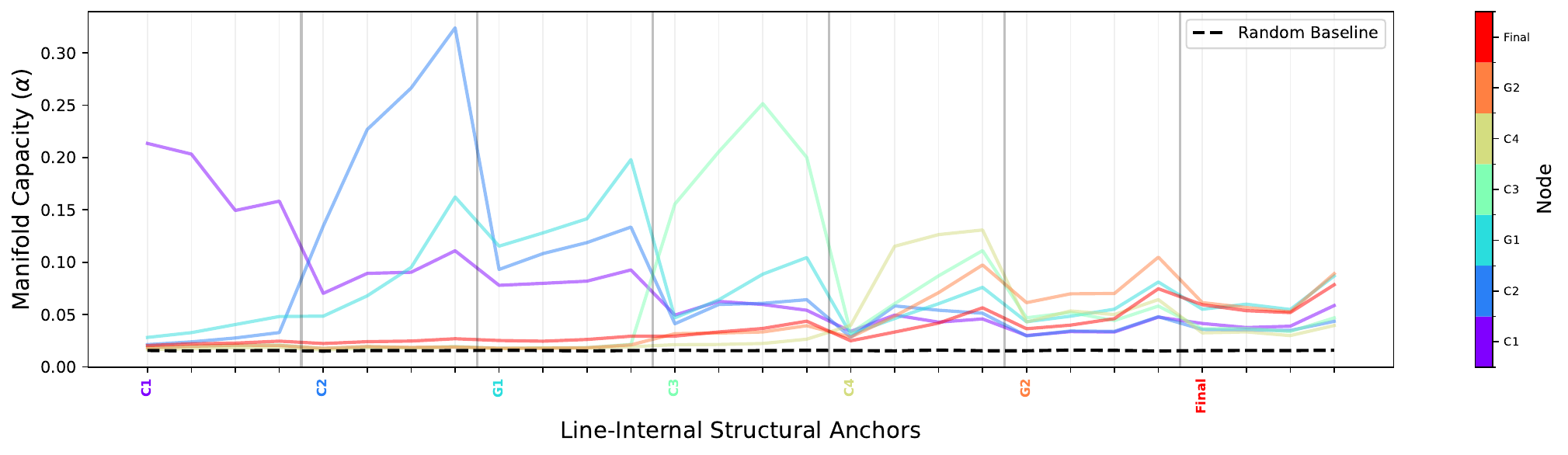}\hfill
  \includegraphics[width=0.48\textwidth]{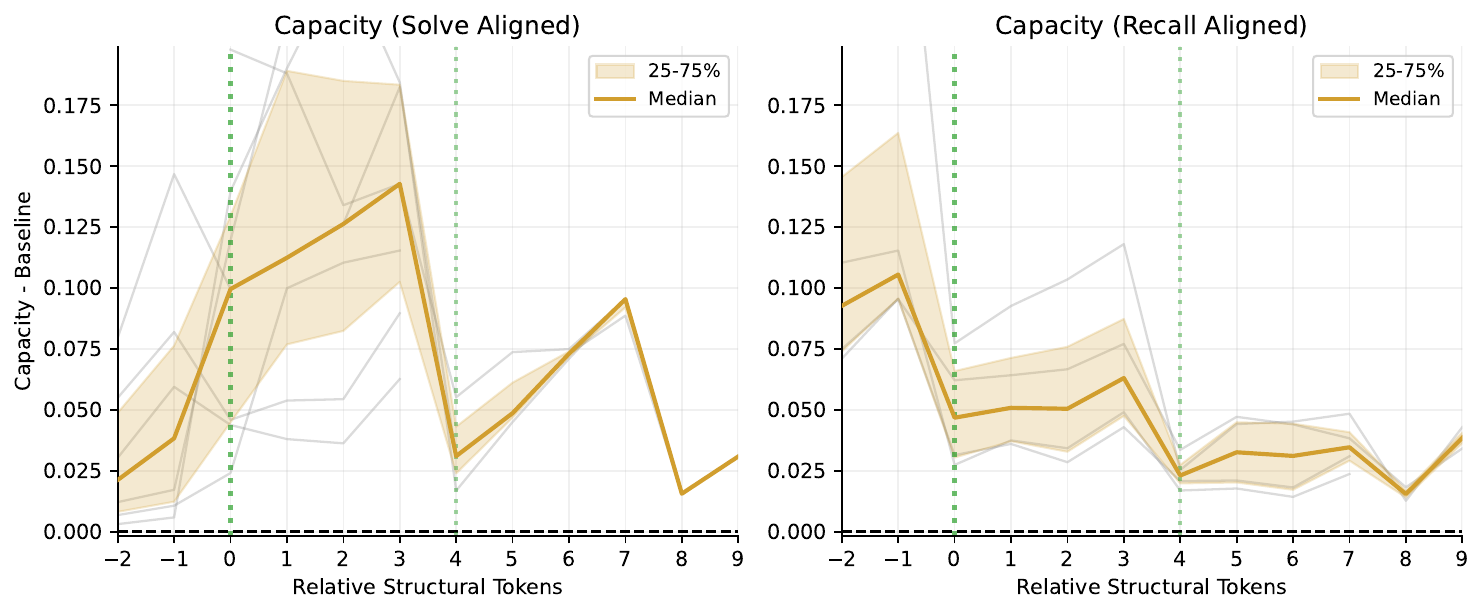}
  \caption{\textbf{Eligibility task, unstructured prompt, Ministral (layer 20).} The model writes free natural-language reasoning lines with only lightweight ID prefixes (\texttt{C1:}, \dots, \texttt{Final:}). Anchors are within-line word-end positions. The pulse and recall pattern is preserved despite the absence of the rigid Markdown template.}
  \label{fig:elig_unstructured}
\end{figure}

\begin{figure}[ht]
  \centering
  \includegraphics[width=0.48\textwidth]{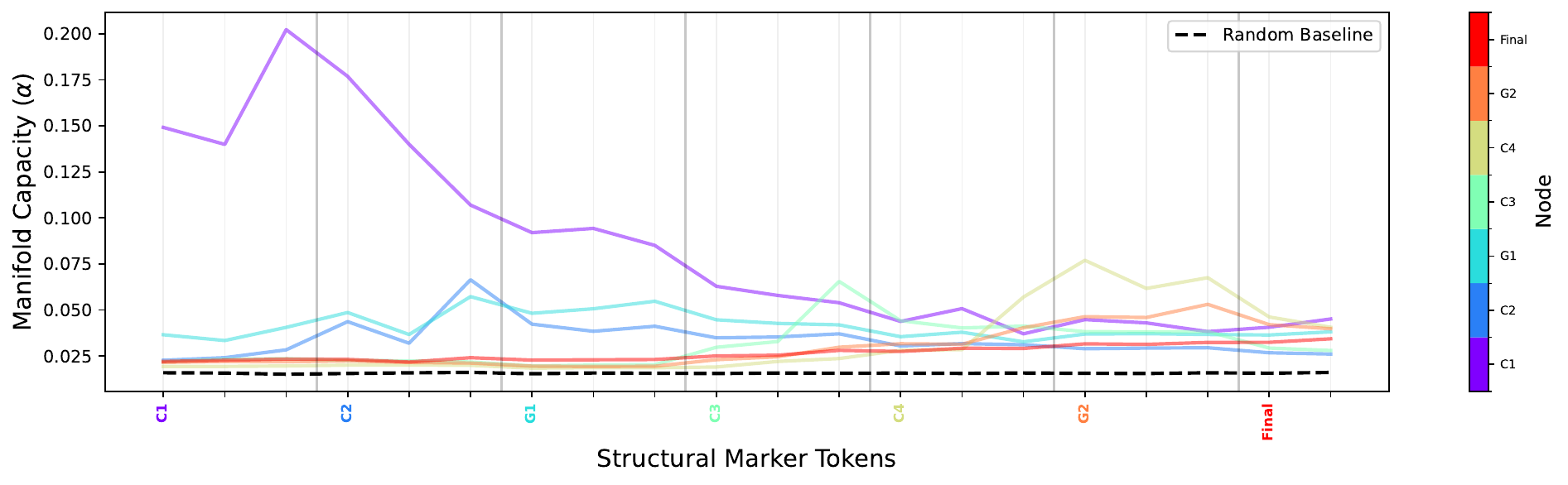}\hfill
  \includegraphics[width=0.48\textwidth]{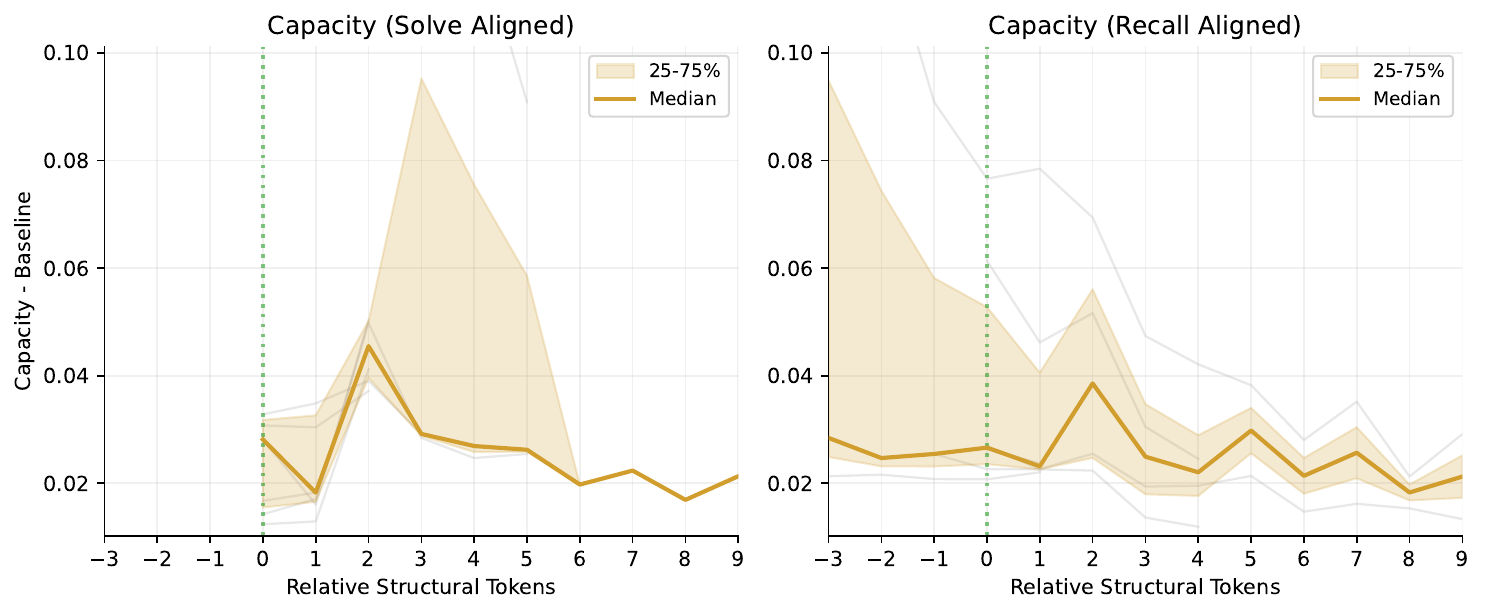}
  \caption{\textbf{Eligibility task, unstructured prompt, Qwen2.5-7B-Instruct (layer 20).} Same conventions as \cref{fig:elig_unstructured}. The pulse and recall pattern is also present on this model.}
  \label{fig:elig_qwen7b_unstructured}
\end{figure}

\section{Probe-vs-Capacity Divergence on the Boolean Task}\label{sec:app_probe_divergence}

\begin{figure}[ht]
  \centering
  \includegraphics[width=0.85\textwidth]{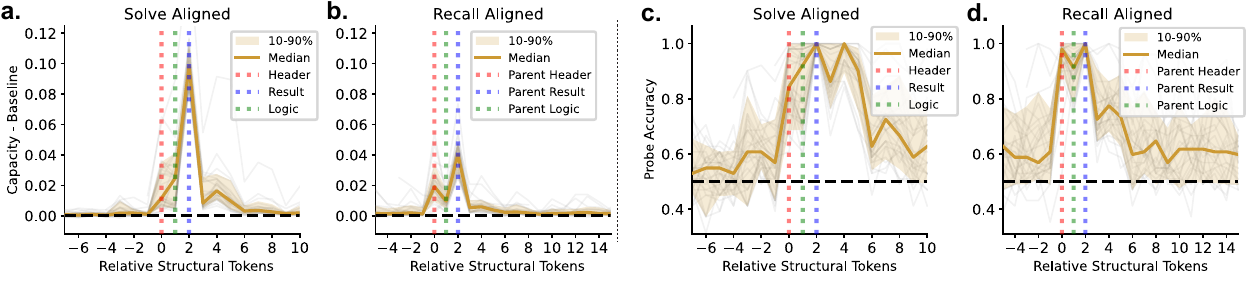}
  \caption{\textbf{Manifold capacity vs.\ hard-margin SVM probe accuracy} (Ministral, Boolean task, layer 20). \textbf{(a, b)} Capacity, aligned to solve and recall (matching panels a and b of \cref{fig:aligned}). \textbf{(c, d)} Hard-margin SVM probe test accuracy at the same anchors: while accuracy also peaks at computation, it exhibits a sustained plateau and decays much more slowly than capacity. A Logistic Regression probe shows the same qualitative behavior (\cref{fig:svm_heatmap}).}
  \label{fig:probe_divergence}
\end{figure}

\Cref{fig:probe_divergence} directly contrasts manifold capacity with hard-margin SVM probe accuracy on the Boolean task. While both signals peak at the moment of computation, the SVM probe accuracy plateaus and decays much more slowly than capacity, supporting the retention-vs-readiness interpretation.

\section{Additional Separability Measures}
\begin{figure}[ht] 
  \begin{center}
    \centerline{\includegraphics[width=0.8\columnwidth]{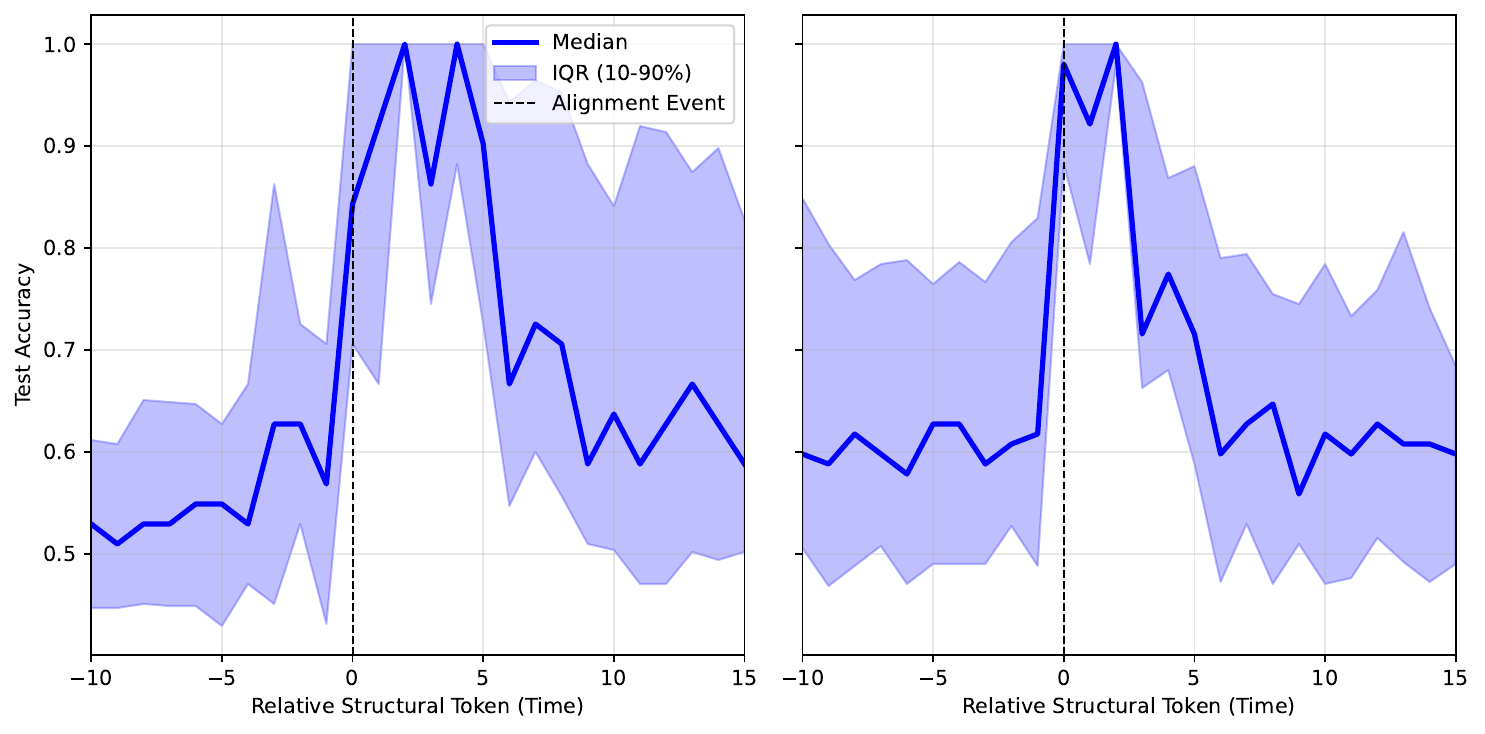}}
    \caption{\textbf{Spatiotemporal heatmap of Logistic Regression probe test accuracy.} (Layer 20) The x-axis represents token positions relative to the computation step ($t=0$). (\emph{During Computation, $t=0,1$}) The test accuracy increase initiates in the middle layers at $t=0$, culminating in a value close to 1 across layers at the token preceding the result ($t=+2$). (\emph{Post-Computation}) For $t > +2$, the early-to-mid layers retain close to 1 accuracy as they retain the immediate context; the accuracy decreases in the middle layers of the model but not very markedly. Only at the subsequent node ($t>+5$) is the decrease more pronounced.
    }
    \label{fig:logistic_traj_SI}
  \end{center}
\end{figure}

\begin{figure}[ht] 
  \centering
  \includegraphics[width=0.48\textwidth]{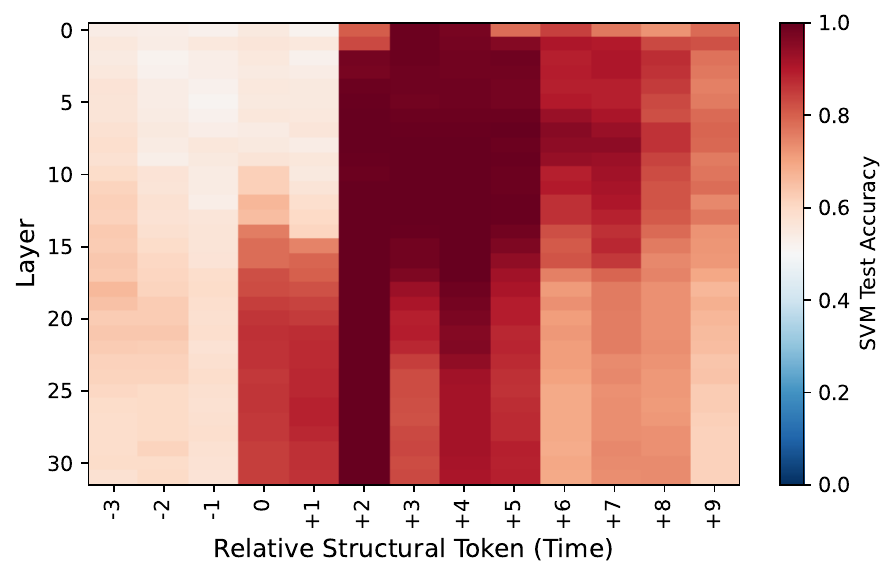}
  \hfill
  \includegraphics[width=0.48\textwidth]{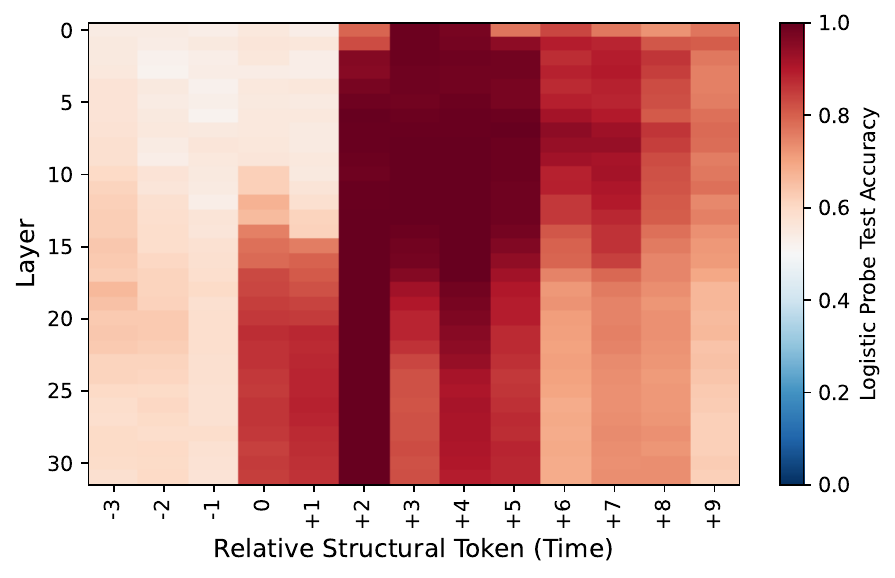}
  
  \caption{\textbf{Spatiotemporal heatmap of Test Accuracy.} \textbf{Left}: Hard-Margin Support Vector Machine. \textbf{Right}: Logistic Probe. The x-axis represents token positions relative to the computation step ($t=0$). (\emph{During Computation, $t=0,1$}) The test accuracy increase initiates in the middle layers at $t=0$, culminating in a value close to 1 across layers at the token preceding the result ($t=+2$). (\emph{Post-Computation}) For $t > +2$, the early-to-mid layers retain close to 1 accuracy as they retain the immediate context; the accuracy decreases in the middle layers of the model but not very markedly. Only at the subsequent node ($t>+5$) is the decrease more pronounced.}
  \label{fig:svm_heatmap}
\end{figure}

Please see Figures \cref{fig:logistic_traj_SI} and \cref{fig:svm_heatmap}.


\section{\emph{Silent} Chain-of-Thought}
\label{sec:silent_cot1}
We analyze the \emph{Silent CoT} setting, where the model is forced to compute the answer without explicitly stating the child nodes' truth values. A representative CoT excerpt is shown in \cref{tab:silent_cot1}.

\begin{table}[ht]
\centering
\small
\begin{tabular}{p{0.3\columnwidth}}
\toprule
\textbf{\emph{Silent} CoT example response excerpt} \\
\midrule
...\\
{*}{*}Node [19]{*}{*}\\
{*} Logic: `[05] and [06]`\\
{*} Result: `True`\\
{*}{*}Node [20]{*}{*}\\
{*} Logic: `[07] xor [08]`\\
{*} Result: `True`\\
...\\
\bottomrule
\end{tabular}
\caption{Excerpt of a Silent CoT response: child-node truth values are not explicitly written in the parent's \texttt{Logic} line.}
\label{tab:silent_cot1}
\end{table}

\begin{figure}[ht] 
  \begin{center}
    \centerline{\includegraphics[width=0.9\columnwidth]{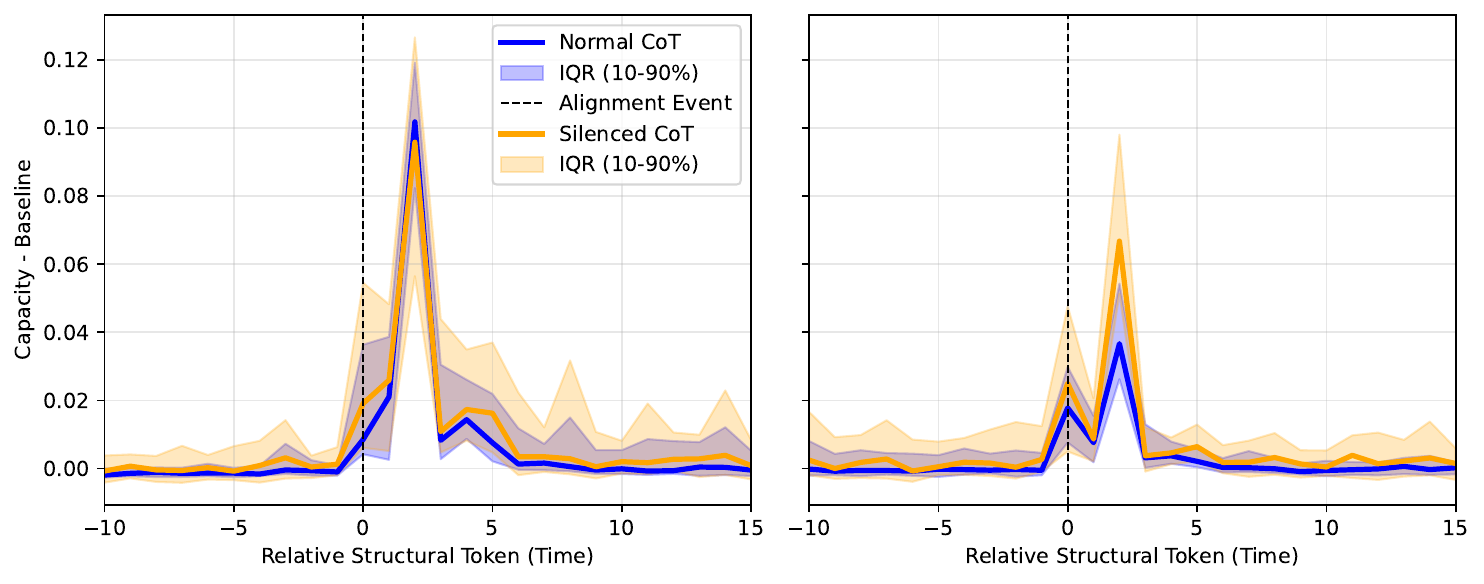}}
    \caption{
    \textbf{Comparison of Capacity dynamics between normal and silenced CoT.} (Left) Data is aligned to the moment a node is computed. There is no statistically significant difference between the two prompts, which is expected as the only difference in the prompts is in the reference to the children.
    (Right) Data is aligned to the moment a node is recalled by its parent. 
    There is no significant difference between the two prompts, which confirms the linear separability of the children is functionally necessary to compute the parent's value and does not merely reflect that the children are being recalled.
    }
    \label{fig:silent_cot1}
  \end{center}
\end{figure}

\Cref{fig:silent_cot1} shows that the manifold capacity of the child node increases during the parent's computation even when the truth values of the child nodes are silenced during the CoT.

In a stronger variant, we prompt Ministral to silence the entire \texttt{* Logic:} line
(\cref{tab:silent_cot2}), replacing a logic expression such as \texttt{[05] and [06]} with \texttt{***************}. This removes the explicit textual content from the parent-computation line while preserving the fixed output format.

\begin{table}[ht]
\centering
\small
\begin{tabular}{p{0.3\columnwidth}}
\toprule
\textbf{Masked-logic CoT example response excerpt} \\
\midrule
...\\
{*}{*}Node [19]{*}{*}\\
{*} Logic: `***************`\\
{*} Result: `True`\\
{*}{*}Node [20]{*}{*}\\
{*} Logic: `***************`\\
{*} Result: `True`\\
...\\
\bottomrule
\end{tabular}
\caption{Excerpt of a masked-logic response: the entire \texttt{Logic} expression is replaced by 15 asterisks of equal token length.}
\label{tab:silent_cot2}
\end{table}

\begin{figure}[ht] 
  \begin{center}
    \centerline{\includegraphics[width=0.75\columnwidth]{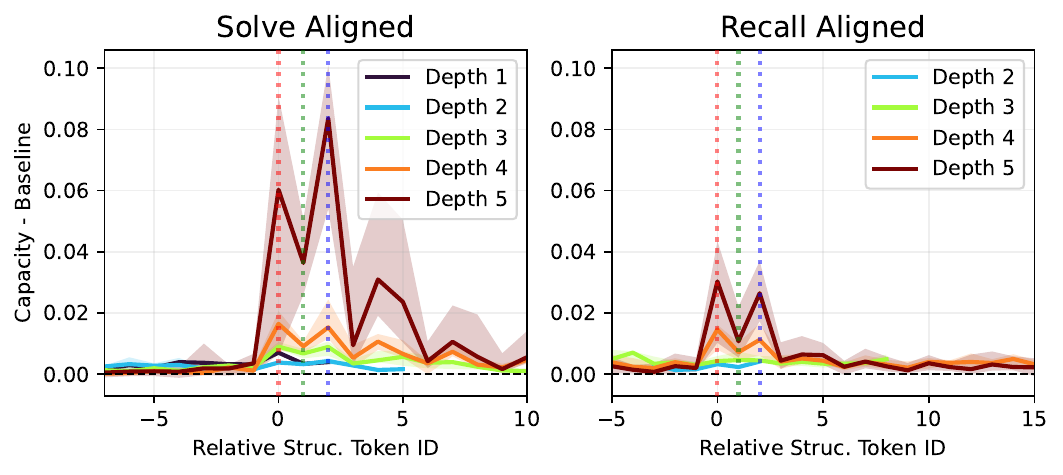}}
    \caption{The masked logic case. Capacity values for each node grouped by the logic tree depth. Left: the node capacity traces aligned at the solve phases of the corresponding nodes. Right: the node capacity traces aligned at the recall phases of the parent nodes (the main focus of this figure). 
    }
    \label{fig:silent_cot2}
  \end{center}
\end{figure}

\begin{figure}[ht] 
  \begin{center}
    \centerline{\includegraphics[width=0.9\columnwidth]{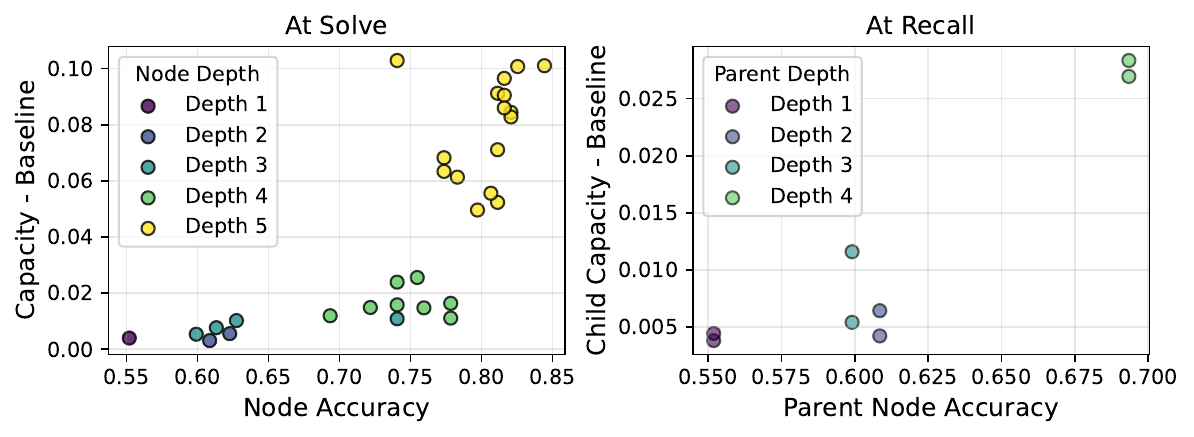}}
    \caption{The masked logic case. Left: node capacity at its solve phase vs. corresponding node accuracy. Right: child node capacity at its recall phase vs. its parent node accuracy. 
    }
    \label{fig:silent_cot2_cap_acc}
  \end{center}
\end{figure}

\begin{figure}[ht] 
  \begin{center}
    \centerline{\includegraphics[width=0.5\columnwidth]{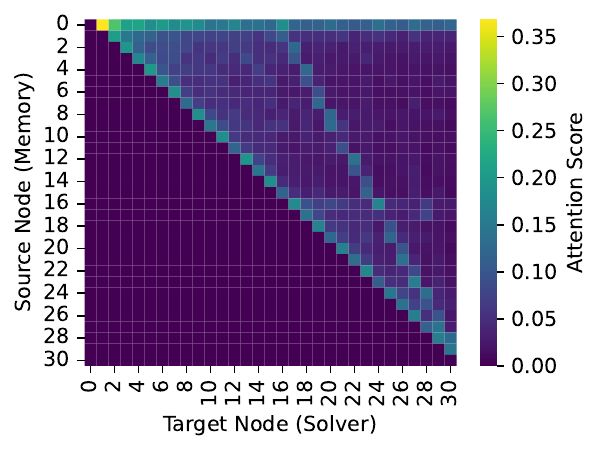}}
    \caption{
    Attention score heatmap between graph nodes for the masked logic case. The x-axis represents the target node currently being solved; the y-axis represents the source nodes in the context. The diagonal structure reflects the model attending to recent context, while off-diagonal points represent the retrieval of specific dependencies.
    }
    \label{fig:silent_cot2_att}
  \end{center}
\end{figure}

The child-node capacity still increases during the solve phase of the parent node, despite the absence of explicit textual output in the logic line (\cref{fig:silent_cot2}). Nodes closer to the root have lower accuracy than nodes closer to the leaves, which naturally leads to lower capacity near the root; this depth trend is visible in \cref{fig:silent_cot2}, where lower depth (e.g.\ depth 2) has lower capacity than higher depth (e.g.\ depth 5). The positive correlation between capacity and accuracy is shown in \cref{fig:silent_cot2_cap_acc}. Finally, the attention map in this setting (\cref{fig:silent_cot2_att}) is qualitatively the same as the standard-prompt attention map (\cref{attention}).

\section{Hyperplane Direction Analysis}\label{sec:supp_direction}
\begin{figure}[ht] 
  \begin{center}
    \centerline{\includegraphics[width=1.0\columnwidth]{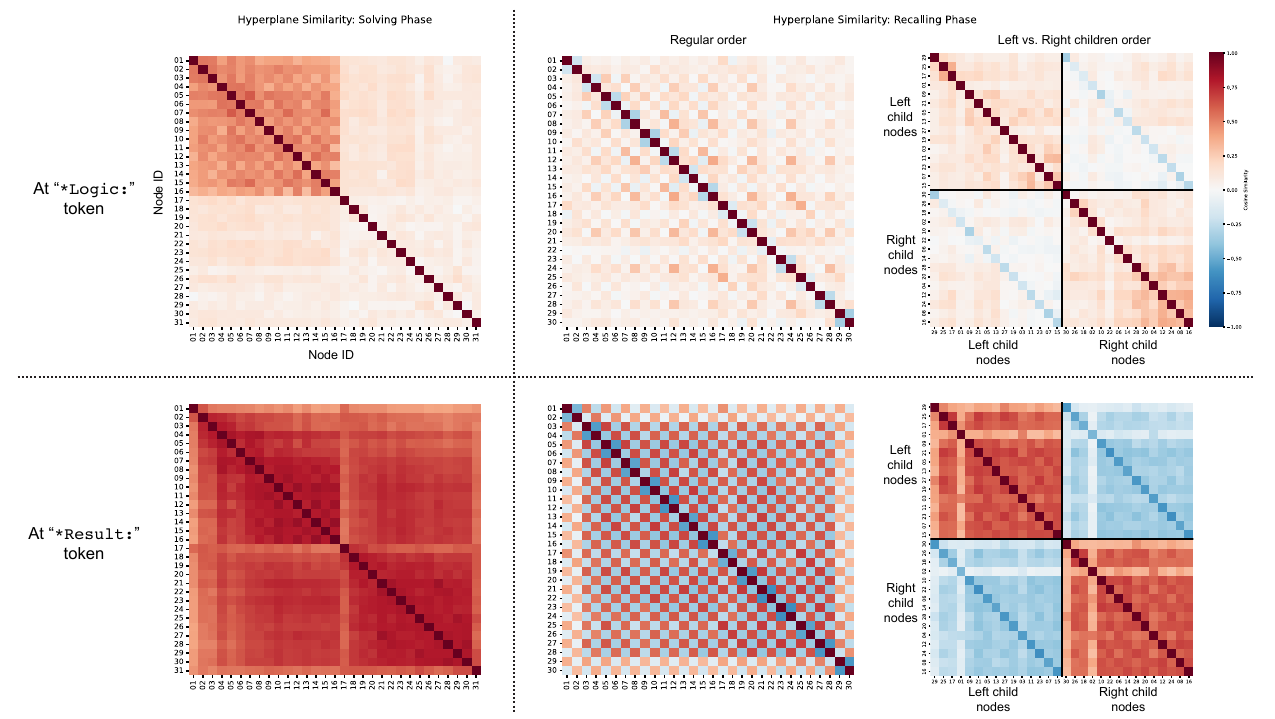}}
    \caption{
    \textbf{Cosine Similarity of Hyperplane Directions between Nodes when They are Solved or Recalled.}
    Cosine similarity between the max-margin hyperplanes computed for nodes' ground-truth answers at the \texttt{* Logic:} tokens (for the \textbf{top} row; \texttt{* Result:} tokens for the \textbf{bottom} row) in their solve phase (for the \textbf{left} column; recall phase for the \textbf{middle} column). For the recall-phase similarity, we also show the same plot with the left and right child nodes grouped, revealing the block structure (the \textbf{right} column).
    }
    \label{fig:inter_node}
  \end{center}
\end{figure}

We compute the max-margin hyperplanes at each token of interest using each node's ground-truth labels, then analyze pairwise cosine similarities between the corresponding normal directions. First, we ask whether all nodes separate along the same direction when they are being solved. At the \texttt{* Logic:} token of the solve phase, only the leaf nodes have highly conserved directions; for the remaining nodes, the directions are mostly orthogonal (\cref{fig:inter_node}, top-left). This suggests that, when a node is actively being solved, each node occupies a distinct direction. At the \texttt{* Result:} token, however, all nodes point toward the same direction, presumably the direction that encodes True or False (\cref{fig:inter_node}, bottom-left). 

We also ask whether the recall direction is conserved across nodes (\cref{fig:inter_node}, middle and right). Consistent with \cref{sec:directions}, the recall directions are highly conserved across all left child nodes and across all right child nodes (\cref{fig:inter_node}, right). Regardless of where in the tree the LLM is solving, it treats the left- and right-child roles consistently. This indicates that the LLM applies the same local rule to compute all subproblems. These two directions (one for the left child and another for the right child) are slightly anti-correlated.

\begin{figure}[ht] 
  \begin{center}
    \centerline{\includegraphics[width=0.6\columnwidth]{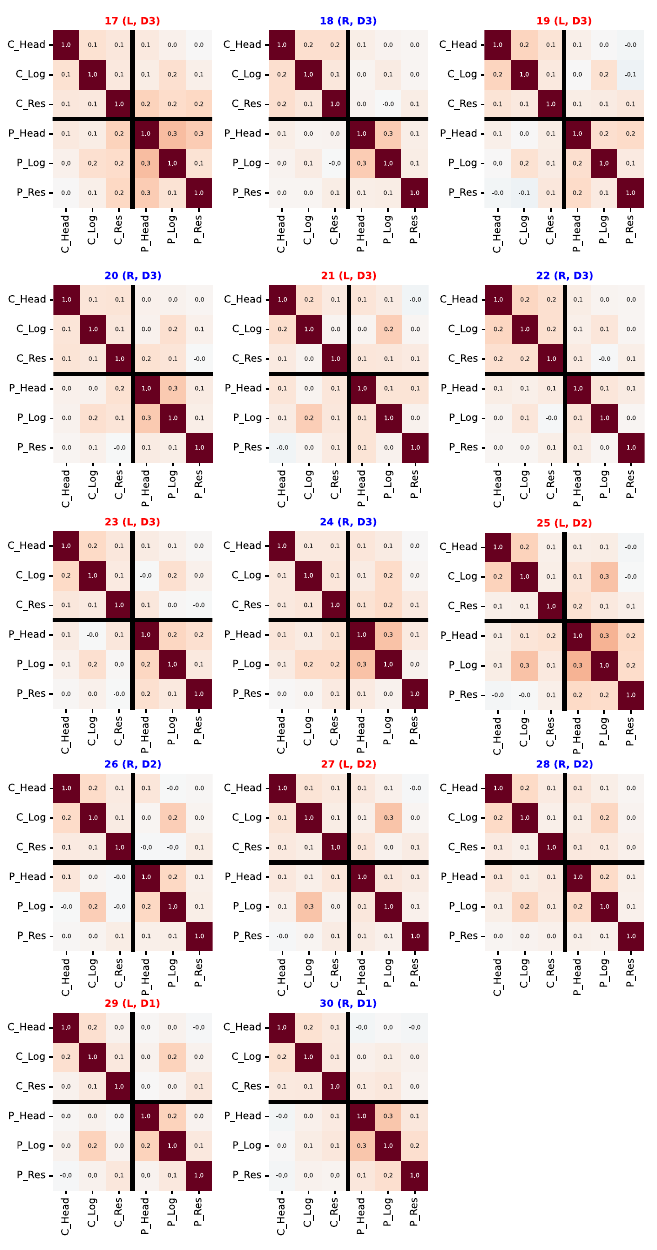}}
    \caption{
    \textbf{Cosine Similarity of Each Node when It is Solved and Recalled.}
    Cosine similarity between the max-margin hyperplanes computed for a single node's ground truth answer when it is being solved (three structural tokens) and recalled (another three structural tokens).
    }
    \label{fig:within_node}
  \end{center}
\end{figure}

Finally, we compare, for each node, the directions used when it is solved and when it is recalled. The solving and recalling directions are mostly orthogonal (\cref{fig:within_node}).

\section{Attention and Manifold Capacity}\label{sec:app_attention}

The recall-time capacity pulse (\cref{fig:aligned}b) raises a natural question: after a node is solved and its capacity decays, where does the information go? If the information is no longer high-capacity in the residual stream, the attention mechanism is the natural candidate for fetching it back from earlier tokens when it is needed again.

\begin{figure}[ht]
  \begin{center}
    \centerline{\includegraphics[width=0.8\columnwidth]{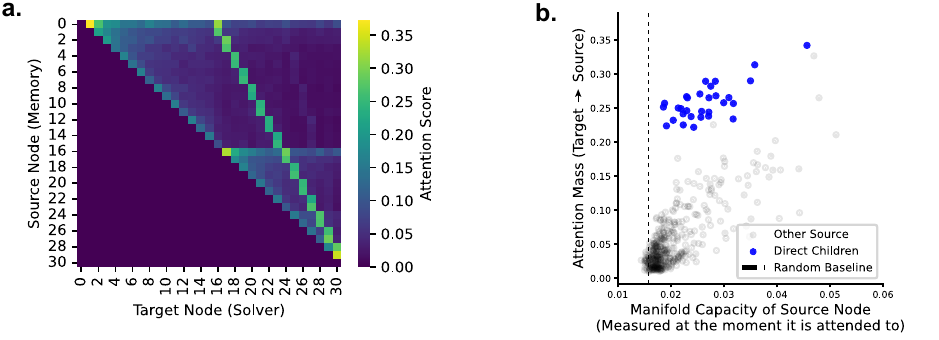}}
    \caption{
    \textbf{Correlation between Attention and Manifold Capacity (Boolean task, Ministral).}
    \textbf{a.} Attention score heatmap between graph nodes. The x-axis represents the target node currently being solved; the y-axis represents the source nodes in the context. The diagonal structure reflects the model attending to recent context, while off-diagonal points represent the retrieval of specific dependencies.
    \textbf{b.} Scatter plot of attention score versus manifold capacity of the source node at the moment of attention. We observe a strong positive correlation ($r=0.723$), specifically for direct children nodes (blue), indicating that the model preferentially attends to previous information when its representation is linearly separable.
    }
    \label{attention}
  \end{center}
\end{figure}

We investigate this in the Boolean setting. \Cref{attention}a shows the attention scores from the target token currently being solved (x-axis) to source tokens corresponding to earlier nodes in the same problem (y-axis); the off-diagonal mass corresponds to the model retrieving specific dependencies, in particular the children of the node currently being solved. \Cref{attention}b correlates the attention score from the target node to a source node with the source node's manifold capacity at that moment: we find a strong positive correlation ($r=0.723$ for direct children, blue). The exact attention quantification is given in \cref{supp:attention_methods}.

These observations suggest that the model preferentially attends to previously visited concepts when their representations are still linearly separable, which is consistent with the view of attention as the mechanism that mediates the dynamic geometric management of the residual stream.

\section{Quantification of Attention Scores}
\label{supp:attention_methods}

To investigate whether the model retrieves information from previous reasoning steps, we analyze the attention patterns during the ``Solve'' phase of each parent node. We define the attention score $A_{T \to S}$ from a target node $T$ (the parent) to a source node $S$ (the child) as follows.

Let $\mathbf{M}^{(l)} \in \mathbb{R}^{H \times N \times N}$ be the multi-head attention matrix at layer $l$, where $H$ is the number of heads and $N$ is the sequence length. We identify the set of token indices corresponding to the source node's solution phase, $\mathcal{I}_S = \{t \mid \text{token}_t \in \text{Result String of } S\}$, and the set of token indices corresponding to the target node's computation phase, $\mathcal{I}_T = \{t \mid \text{token}_t \in \text{Logic String of } T\}$.

We compute the maximum attention score across all heads and all token pairs within the defined windows:
\begin{equation}
    A_{T \to S} = \max_{l \in L_\text{ROI}} \left[ \max_{h \in \{1 \dots H\}} \left( \max_{i \in \mathcal{I}_T, j \in \mathcal{I}_S} \mathbf{M}^{(l)}_{h, i, j} \right) \right]
\end{equation}
where $L_\text{ROI}$ represents the layers of interest (e.g., Layer 20). The maximum aggregation is chosen over the mean to detect sparse retrieval mechanisms, as information transfer in transformers is often performed by specialized induction heads rather than distributed uniformly across all heads.

For the correlation analysis shown in Figure \ref{attention}b, we align this attention score $A_{T \to S}$ with the manifold capacity of the source node $S$, $\alpha^{(S)}$, measured using the embeddings at the last token of the target phase $\mathcal{I}_T$. This allows us to directly compare the strength of retrieval with the resulting linear separability of the retrieved concept.

\section{Additional Capacity and Probe Traces}
\begin{figure}[ht] 
  \begin{center}
    \centerline{\includegraphics[width=0.8\columnwidth]{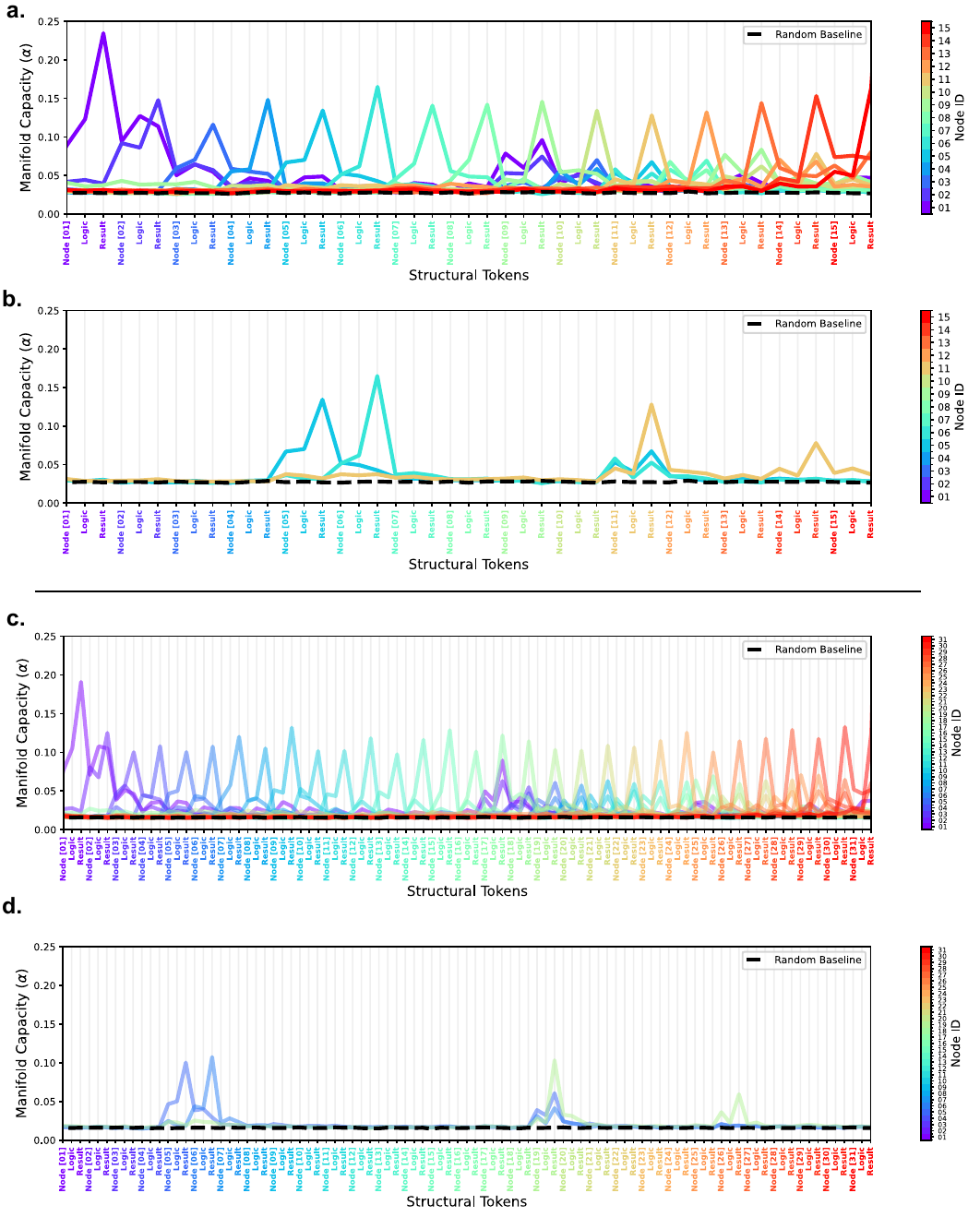}}
    \caption{
    \textbf{Dynamic Modulation of Manifold Geometry during CoT.}
    \textbf{a.} Manifold capacity ($\alpha$) tracked across the Chain-of-Thought sequence (layer 20, tree height $h=4$). Lines are colored by node ID. Capacity for a specific node peaks sharply at two distinct moments: first when the node is intrinsically computed, and second when it is processed by its parent.
    \textbf{b.} Detailed analysis of Node 11 (yellow) and its children, Node 5 (blue) and Node 6 (turquoise). High capacity peaks correspond to linearly separable manifolds (clear decision boundaries). Notably, representations become entangled (low capacity) in the interim periods between the initial solution (solve) and the subsequent retrieval (recall). 
    \textbf{c, d.} Same views for a tree of height $h=5$.
    }
    \label{fig:capacity_trace_SI}
  \end{center}
\end{figure}

\begin{figure}[ht] 
  \begin{center}
    \centerline{\includegraphics[width=0.8\columnwidth]{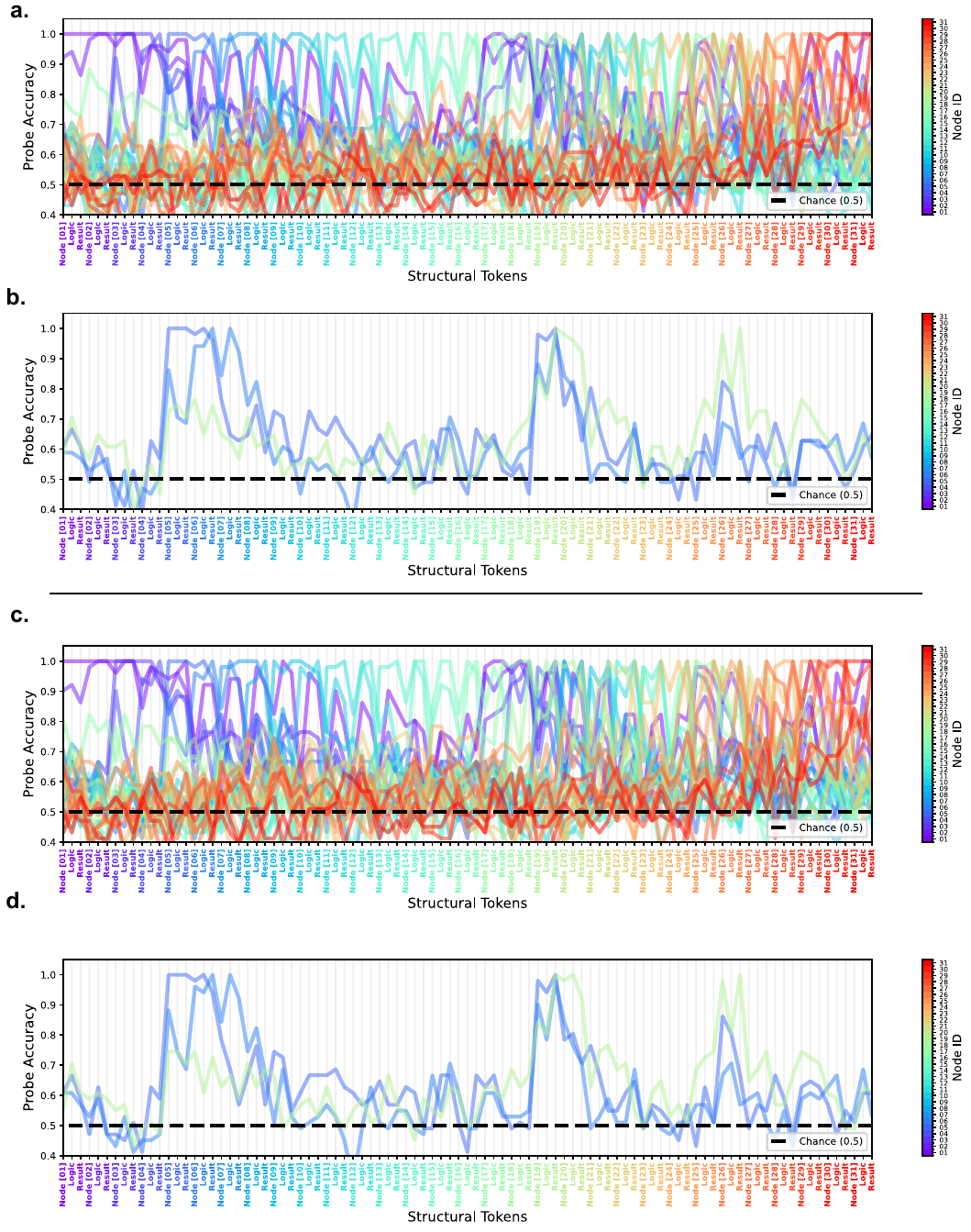}}
    \caption{
    \textbf{Evolution of Linear Probe Test Accuracy during CoT.}
    \textbf{a.} Logistic Probe Test Accuracy across the Chain-of-Thought sequence (layer 20, tree height $h=5$). Lines are colored by node ID. 
    \textbf{b.} Detailed analysis of Node 11 (yellow) and its children, Node 5 (blue) and Node 6 (turquoise). Accuracy peaks when each node is being solved and when it is being recalled. However, unlike manifold capacity (see panels c and d of \cref{fig:capacity_trace_SI}), accuracy plateaus for several nodes.
    \textbf{c, d.} Same views for hard-margin SVM probe test accuracy.
    }
    \label{fig:capacity_probe_SI}
  \end{center}
\end{figure}

Additional capacity and probe traces are shown in \cref{fig:capacity_trace_SI,fig:capacity_probe_SI}.

\end{document}